\documentclass[lettersize,journal]{IEEEtran}
\usepackage{amsmath,amsfonts}
\usepackage{algorithmic}
\usepackage{algorithm}
\usepackage{array}
\usepackage[caption=false,font=normalsize,labelfont=sf,textfont=sf]{subfig}
\usepackage{textcomp}
\usepackage{stfloats}
\usepackage{url}
\usepackage{verbatim}
\usepackage{graphicx}
\usepackage{cite}
\usepackage{booktabs}

\usepackage{xcolor}
\usepackage{colortbl}
\usepackage{tikz}
\usepackage{arydshln}

\definecolor{myblue}{RGB}{0,114,189}
\newcommand{\blue}{\cellcolor{myblue!10}}

\definecolor{taxBlue}{RGB}{82,135,235}
\definecolor{taxBlueFill}{RGB}{224,235,250}
\definecolor{memGreen}{RGB}{79,157,91}
\definecolor{memGreenFill}{RGB}{224,241,227}
\definecolor{expOrange}{RGB}{245,135,69}
\definecolor{expOrangeFill}{RGB}{255,247,219}
\definecolor{evalRed}{RGB}{235,86,86}
\definecolor{evalRedFill}{RGB}{253,232,230}
\definecolor{futurePurple}{RGB}{178,86,220}
\definecolor{futurePurpleFill}{RGB}{244,235,252}
\definecolor{lineGray}{RGB}{170,170,170}

\begin{document}

\title{Memory for Large Language Models}

\author{
  Sining Zhoubian,
  Dan Zhang,
  Evgeny Kharlamov,
  and Jie Tang,~\IEEEmembership{Fellow, IEEE}
\thanks{Sining Zhoubian, Tsinghua University, zbsn21@tsinghua.mails.edu.cn.}
\thanks{Dan Zhang, National University of Singapore, zhangdan25@nus.edu.sg.}
\thanks{Evgeny Kharlamov, Bosch AI, evgeny.kharlamov@de.bosch.com.}
\thanks{Jie Tang, Tsinghua University, jietang@tsinghua.edu.cn.}
}


\markboth{Preprint}%
{Zhoubian \MakeLowercase{\textit{et al.}}: Memory for Large Language Models}


\maketitle

\begin{abstract}
  Memory has evolved into a foundational architectural dimension in large language models (LLMs), shifting from an implicit byproduct of computation to a spectrum of explicit, controllable mechanisms. 
  While recent advances introduce diverse strategies---spanning transient attention, recurrent state dynamics, parameter-efficient adaptations, and scalable lookup storage---this rapid evolution has led to a highly fragmented research landscape. 
  In this survey, we present a systematic, architecture-centric taxonomy of memory in LLMs. 
  Our framework characterizes memory along three orthogonal axes: representation (implicit versus explicit), update dynamics (offline versus online), and persistence (short-term versus long-term). 
  We further formalize the granular mechanisms dictating memory writing, routing, state transitions, and consolidation. 
  This unified perspective elucidates the conceptual boundaries between computation-coupled and independently addressable memory, effectively bridging disparate architectural paradigms. 
  Additionally, we critically analyze hybrid memory architectures, system-level efficiency trade-offs, and multi-dimensional evaluation methodologies. 
  By consolidating these scattered advancements into a cohesive framework, this survey charts the trajectory of memory-centric LLM design and provides a principled foundation for future innovations in scalable and adaptive language modeling.
\end{abstract}

\begin{IEEEkeywords}
Large Language Models, Memory, Architecture, Data Compression.
\end{IEEEkeywords}

\section{Introduction}

\IEEEPARstart{L}{arge} language models (LLMs) have demonstrated remarkable capabilities across reasoning, coding, dialogue, and long-context understanding. 
While much of their success is attributed to scaling laws---encompassing parameters, data, and compute---an equally fundamental architectural dimension is increasingly evident: \emph{memory}. 
From the transient attention mechanisms that mediate token-level interactions to recurrent state dynamics and persistent test-time parameter adaptations, modern LLMs increasingly rely on multifaceted memory systems to retain, retrieve, and update information across extended sequences and interactions.
Historically, \emph{memory in neural language models was predominantly implicit}.
Transformer attention provided a content-addressable working memory bounded by a fixed context window, enabling flexible short-term information aggregation. 
However, its computational and storage overhead scales quadratically with sequence length. 
Subsequent advances extended this paradigm through sparse attention \cite{child2019sparse,zaheer2020bigbird,beltagy2020longformer}, structured state-space models \cite{gu2024mamba,dao2024mamba2}, and hybrid architectures coupling recurrent and attentional dynamics \cite{lieber2024jambahybridtransformermambalanguage,jambateam2024jamba15hybridtransformermambamodels}. 
While these developments significantly extended accessible context lengths and enhanced computational efficiency, they often introduced performance trade-offs. 
Furthermore, memory within these architectures remains inherently ephemeral and tightly coupled to the forward computational graph.

\begin{figure*}[t!]
    \centering
    \includegraphics[width=0.9\linewidth]{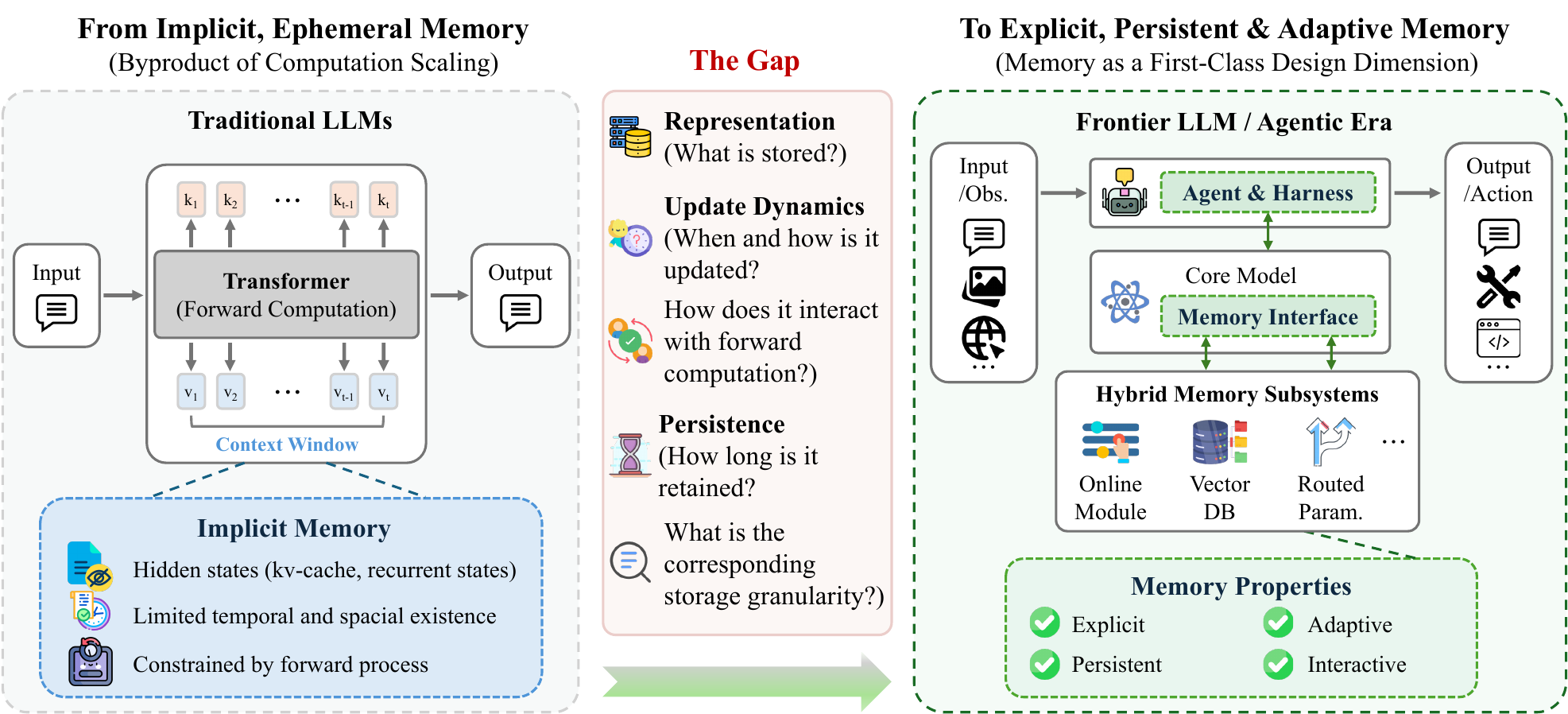}
    \caption{The evolution of memory in LLMs. As memory transitions from an implicit byproduct of computation to an explicit architectural design dimension in modern LLMs and AI agents, diverse memory mechanisms have emerged with different storage forms, persistence properties, and update dynamics. This conceptual shift motivates a unified taxonomy for characterizing and comparing memory architectures.}
    \label{fig:transfer}
\end{figure*}

More recently, a distinct architectural shift has emerged. 
As illustrated in Figure~\ref{fig:transfer}, an increasing number of studies incorporate \emph{explicit and persistent memory mechanisms} that adaptively evolve beyond ephemeral inference, providing models with orthogonal avenues for capability expansion. 
Approaches such as Titans \cite{behrouz2025titans} and end-to-end test-time training \cite{tandon2025endtoend} enable dynamic parameter updates during deployment.
Lookup-based architectures, including Engram \cite{cheng2026conditionalmemoryscalablelookup} decouple memory storage from dense computation. 
Nested and multi-timescale update strategies \cite{behrouz2025nested} further blur the boundary between conventional training and inference. 
Concurrently, conditional parameter modules and mixture-of-experts (MoE) architectures~\cite{deepseek2024moe,mixtral2024} selectively activate parameter subsets based on input context, constituting a structured form of conditional memory. 
Collectively, these advancements signal that memory is transitioning into a primary, explicit architectural design dimension rather than remaining a mere byproduct of scaling.

Despite this rapid progress, the literature lacks a unified perspective elucidating the conceptual relationships among these disparate memory mechanisms. 
Attention-based caching, recurrent sequence states, test-time adaptation, retrieval modules, and conditional parameter routing are frequently studied in isolation, even though they address fundamentally isomorphic questions: \emph{what is stored, when and how it is updated, how it interacts with forward computation, its retention duration, and its underlying storage granularity}. 
Without a principled structural framework, comparing methodologies, analyzing system-level trade-offs, and identifying unexplored design spaces remain profoundly difficult.

\begin{figure*}[t!]
    \centering
    \resizebox{\textwidth}{!}{
        \begin{tikzpicture}[
  x=1cm,
  y=1cm,
  font=\sffamily,
  connector/.style={draw=lineGray, line width=0.65pt},
  root/.style={
    draw=lineGray,
    fill=gray!10,
    rounded corners=2pt,
    minimum width=1.10cm,
    minimum height=1.85cm,
    align=center,
    rotate=90,
    font=\sffamily\Large
  },
  levelone/.style={
    rounded corners=2pt,
    minimum width=4.25cm,
    minimum height=1.18cm,
    align=center,
    font=\sffamily\Large
  },
  midbox/.style={
    rounded corners=2pt,
    minimum width=4.85cm,
    minimum height=1.14cm,
    align=center,
    font=\sffamily\large
  },
  descbox/.style={
    rounded corners=2pt,
    minimum width=10.20cm,
    minimum height=1.25cm,
    text width=9.55cm,
    align=left,
    font=\rmfamily\normalsize
  }
]

\coordinate (rootC) at (0,11.8);
\coordinate (l1x) at (3.8,0);
\coordinate (midx) at (9.1,0);
\coordinate (descx) at (17.1,0);

\node[root] (root) at (rootC) {Memory \\ for LLMs};
\draw[connector] (0.62,12.0) -- (1.10,12.0) coordinate (trunk);
\draw[connector] (trunk) |- (1.635,23.0);
\draw[connector] (trunk) |- (1.635,18.2);
\draw[connector] (trunk) |- (1.635,12.0);
\draw[connector] (trunk) |- (1.635,5.6);
\draw[connector] (trunk) |- (1.635,1.45);

\node[levelone, draw=taxBlue, fill=taxBlueFill] (tax) at (3.8,23.0) {Taxonomy (\S\ref{sec:taxonomy})};

\node[midbox, draw=taxBlue, fill=taxBlueFill] (rep) at (9.1,24.4) {Representation (\S\ref{subsec:representation})};
\node[midbox, draw=taxBlue, fill=taxBlueFill] (upd) at (9.1,23.0) {Update Dynamics (\S\ref{subsec:update})};
\node[midbox, draw=taxBlue, fill=taxBlueFill] (per) at (9.1,21.6) {Persistence (\S\ref{subsec:persistence})};

\node[descbox, draw=taxBlue] (repd) at (17.1,24.4) {
  $\bullet$ Implicit: Compute-coupled (KV Cache, Hidden States)\\
  $\bullet$ Explicit: Decoupled interface (Datastore, Parameter modules)
};
\node[descbox, draw=taxBlue] (updd) at (17.1,23.0) {
  $\bullet$ Offline: Updates during training only (Pretraining, MoE)\\
  $\bullet$ Online: Updates during inference (TTT, Recurrent States)
};
\node[descbox, draw=taxBlue] (perd) at (17.1,21.6) {
  $\bullet$ Short-Term: Transient, bounded to context window (Attention Cache)\\
  $\bullet$ Long-Term: Persistent across contexts/sessions (Titans, Engram)
};

\draw[connector] (tax.east) -- ++(0.31,0) coordinate (taxJoin);
\draw[connector] (taxJoin) |- (rep.west);
\draw[connector] (taxJoin) -- (upd.west);
\draw[connector] (taxJoin) |- (per.west);
\draw[connector] (rep.east) -- (repd.west);
\draw[connector] (upd.east) -- (updd.west);
\draw[connector] (per.east) -- (perd.west);

\node[levelone, draw=memGreen, fill=memGreenFill] (imp) at (3.8,18.2) {Implicit Memory (\S\ref{sec:implicit})};

\node[midbox, draw=memGreen, fill=memGreenFill] (attn) at (9.1,19.6) {Attention-based (\S\ref{subsec:attention_memory})};
\node[midbox, draw=memGreen, fill=memGreenFill] (sparse) at (9.1,18.2) {Sparse \& Selective (\S\ref{subsec:sparse_memory})};
\node[midbox, draw=memGreen, fill=memGreenFill] (ssm) at (9.1,16.8) {Recurrent (\S\ref{subsec:ssm_memory})};

\node[descbox, draw=memGreen] (attnd) at (17.1,19.6) {
  e.g., Transformer (KV Cache), StreamingLLM, HyperMLP,\\
  Sliding-Window and Dynamic Sequence-Mixing Attention
};
\node[descbox, draw=memGreen] (sparsed) at (17.1,18.2) {
  e.g., Selective Attention, RATTENTION, MoBA, NSA,\\
  Content-Routed and Local--Global KV Access
};
\node[descbox, draw=memGreen] (ssmd) at (17.1,16.8) {
  e.g., Mamba-3, RWKV-7, Gated DeltaNet-2,\\
  Log-Linear Attention, Kalman Linear Attention
};

\draw[connector] (imp.east) -- ++(0.15,0) coordinate (impJoin);
\draw[connector] (impJoin) |- (attn.west);
\draw[connector] (impJoin) -- (sparse.west);
\draw[connector] (impJoin) |- (ssm.west);
\draw[connector] (attn.east) -- (attnd.west);
\draw[connector] (sparse.east) -- (sparsed.west);
\draw[connector] (ssm.east) -- (ssmd.west);

\node[levelone, draw=expOrange, fill=expOrangeFill] (exp) at (3.8,12.0) {Explicit Memory (\S\ref{sec:explicit})};

\node[midbox, draw=expOrange, fill=expOrangeFill] (param) at (9.1,14.1) {Parameterized (\S\ref{subsec:parameterized})};
\node[midbox, draw=expOrange, fill=expOrangeFill] (lookup) at (9.1,12.7) {Lookup-based (\S\ref{subsec:lookup})};
\node[midbox, draw=expOrange, fill=expOrangeFill] (moe) at (9.1,11.3) {MoE (\S\ref{subsec:moe_memory})};
\node[midbox, draw=expOrange, fill=expOrangeFill] (multi) at (9.1,9.9) {Multi-Timescale (\S\ref{subsec:multi-timescale})};

\node[descbox, draw=expOrange] (paramd) at (17.1,14.1) {
  e.g., Titans, TTT-E2E, MEMORYLLM, LM2,\\
  In-Place TTT, and Dedicated Writable Modules
};
\node[descbox, draw=expOrange] (lookupd) at (17.1,12.7) {
  e.g., kNN-LM, PlugLM, Engram, ExplicitLM,\\
  Editable Datastores, and Hashed Lookup Slots
};
\node[descbox, draw=expOrange] (moed) at (17.1,11.3) {
  e.g., Switch Transformer, Mixtral, DeepSeek-MoE,\\
  Routed Expert Subnetworks as Conditional Memory
};
\node[descbox, draw=expOrange] (multid) at (17.1,9.9) {
  e.g., Nested Learning (Hope), Slow--Fast Modules,\\
  Continuum Memory, and Hierarchical Update Schedules
};

\draw[connector] (exp.east) -- ++(0.21,0) coordinate (expJoin);
\draw[connector] (expJoin) |- (param.west);
\draw[connector] (expJoin) |- (lookup.west);
\draw[connector] (expJoin) |- (moe.west);
\draw[connector] (expJoin) |- (multi.west);
\draw[connector] (param.east) -- (paramd.west);
\draw[connector] (lookup.east) -- (lookupd.west);
\draw[connector] (moe.east) -- (moed.west);
\draw[connector] (multi.east) -- (multid.west);

\node[levelone, draw=evalRed, fill=evalRedFill] (sys) at (3.8,5.6) {System \& Eval (\S\ref{sec:systems})};

\node[midbox, draw=evalRed, fill=evalRedFill] (hybrid) at (9.1,7.0) {Hybrid Architectures (\S\ref{subsec:hybrid_systems})};
\node[midbox, draw=evalRed, fill=evalRedFill] (eff) at (9.1,5.6) {Efficiency \& Mgmt (\S\ref{subsec:memory_management})};
\node[midbox, draw=evalRed, fill=evalRedFill] (eval) at (9.1,4.2) {Evaluation (\S\ref{subsec:evaluation})};

\node[descbox, draw=evalRed] (hybridd) at (17.1,7.0) {
  e.g., Jamba, Kimi Linear, OLMo Hybrid, AMOR, HAM,\\
  Fixed and Adaptively Routed Complementary Memories
};
\node[descbox, draw=evalRed] (effd) at (17.1,5.6) {
  e.g., CommVQ, PagedAttention, Memory Caching,\\
  Bottlenecked Transformers, and KV Consolidation
};
\node[descbox, draw=evalRed] (evald) at (17.1,4.2) {
  e.g., RULER, LongBench, $\infty$Bench, L-Eval, SCROLLS,\\
  Recall, Structured Reasoning, and Efficiency Diagnostics
};

\draw[connector] (sys.east) -- ++(0.25,0) coordinate (sysJoin);
\draw[connector] (sysJoin) |- (hybrid.west);
\draw[connector] (sysJoin) -- (eff.west);
\draw[connector] (sysJoin) |- (eval.west);
\draw[connector] (hybrid.east) -- (hybridd.west);
\draw[connector] (eff.east) -- (effd.west);
\draw[connector] (eval.east) -- (evald.west);

\node[levelone, draw=futurePurple, fill=futurePurpleFill] (future) at (3.8,1.45) {Future Directions (\S\ref{sec:challenges})};
\node[midbox, draw=futurePurple, fill=futurePurpleFill] (open) at (9.1,1.45) {Open Challenges};
\node[descbox, draw=futurePurple, minimum height=1.45cm] (opend) at (17.1,1.45) {
  $\bullet$ Unified Theory of Memory $\bullet$ Lifelong Parametric Memory\\
  $\bullet$ Adaptive Memory Allocation $\bullet$ Hardware-Algorithm Co-Design\\
  $\bullet$ Multi-Dimensional Evaluation Framework
};

\draw[connector] (future.east) -- (open.west);
\draw[connector] (open.east) -- (opend.west);

\end{tikzpicture}
    }
    \vspace{-4mm}
    \caption{Main content and taxonomy of memory for LLMs.}
    \label{fig:content}
\end{figure*}
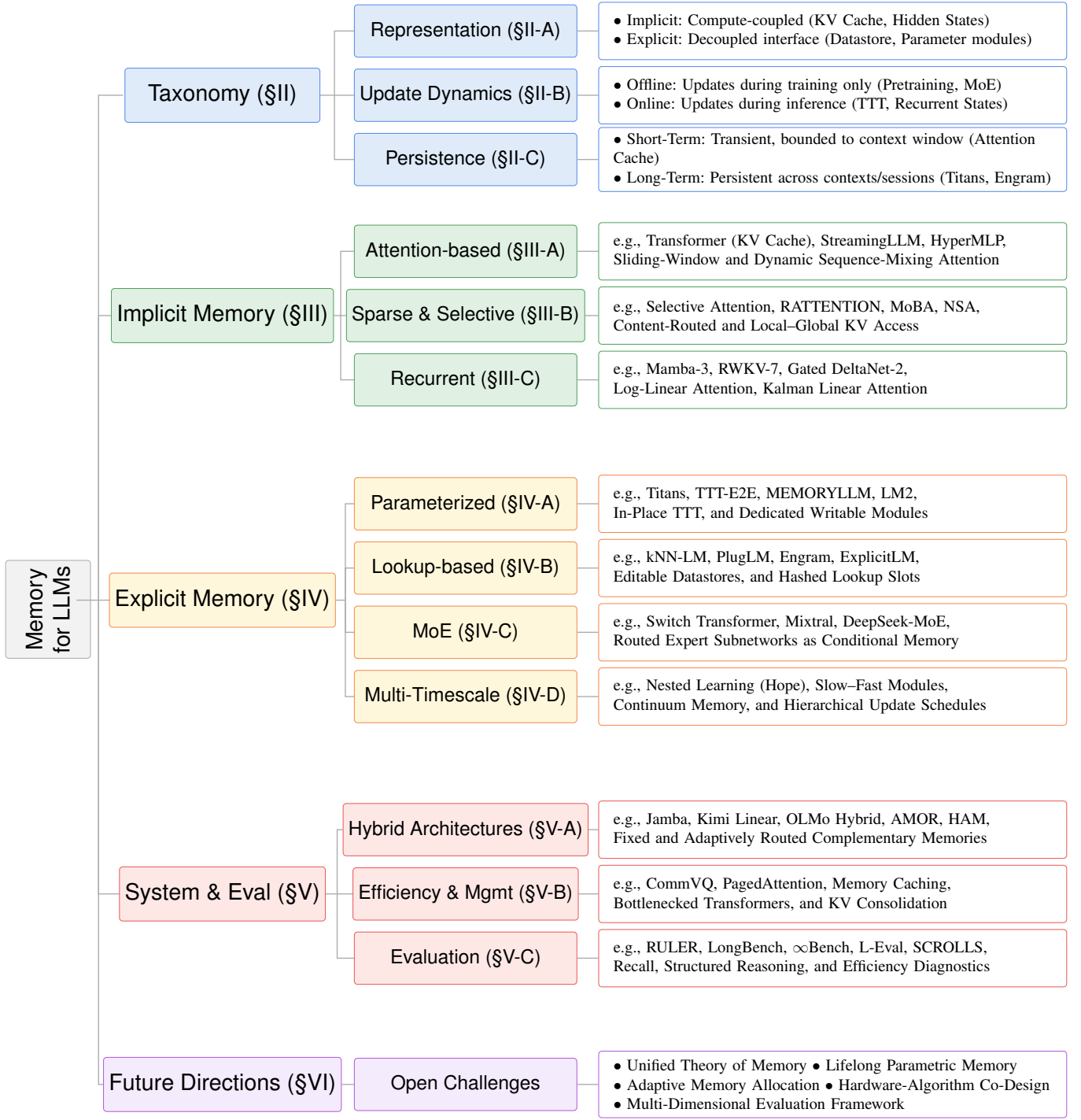

In this survey, we present a systematic, architecture-centric treatment of memory in LLMs. 
We introduce a comprehensive taxonomy characterizing memory along three orthogonal axes: 
1) representation form, distinguishing implicit computational memory from explicit parameterized or structured storage; 
2) persistence, capturing the temporal lifespan of retained information; and 
3) update dynamics, describing when and how memory representations evolves. 
This taxonomy facilitates a coherent interpretation of a broad spectrum of recent architectures, clarifying their structural and conceptual interdependencies. 
By reframing these scattered advancements through a memory-centric lens, this survey aims to consolidate an emerging research frontier and provide a principled foundation for future architectural innovations. 
In Figure~\ref{fig:content}, we summarize the core content and taxonomic structure of this survey.

Our primary contributions are threefold:
\begin{itemize}
    \item We propose a unified memory taxonomy for LLM architectures, establishing precise terminology to delineate implicit, explicit, persistent, and adaptive memory mechanisms (cf. Section~\ref{sec:taxonomy}).
    \item We systematically review recent advances across computation-driven and parameterized memory systems, encompassing hybrid architectures, multi-timescale update policies, and conditional parameter routing (cf. Section~\ref{sec:implicit} and Section~\ref{sec:explicit}).
    \item We critically analyze system-level trade-offs and multi-dimensional evaluation methodologies for memory-centric models, highlighting persistent open challenges and promising future research trajectories (cf. Section~\ref{sec:systems} and Section~\ref{sec:challenges}).
\end{itemize}

\section{A Taxonomy of Memory in Large Language Models}
\label{sec:taxonomy}

Despite the growing body of work on memory-augmented large language models, the term ``memory'' is often used ambiguously to refer to heterogeneous mechanisms or systems operating at different levels of abstraction. In this survey, we focus specifically on \emph{model-level memory mechanisms and systems}, i.e., memory that is instantiated within the model architecture or its inference-time dynamics, and exclude agent-level or prompt-based memory systems that rely on external orchestration or engineering pipelines. To provide a unified and principled view, we organize existing approaches along three orthogonal axes: \emph{representation}, \emph{update dynamics}, and \emph{persistence}. These axes capture what is stored as memory, when and how it is updated, and how long it persists, respectively. Importantly, our taxonomy is not intended to be a rigid classification, but rather a conceptual framework that reveals common structures underlying seemingly diverse methods. 
In Table~\ref{tab:memory_taxonomy}, we present a summary of recent work concerning LLM memory systems, categorized across the three axes.

\begin{table*}[!t]
\centering
\caption{A summary of representative research concerning modern LLM memory systems based on the proposed taxonomy. The table categorizes models across three axes: Representation, Update Dynamics, and Persistence. Note that all models are represented by their primary mechanism of memory, and some models may simultaneously encompass multiple features within the same axis.}
\label{tab:memory_taxonomy}
\small
\resizebox{\textwidth}{!}{
    \begin{tabular}{@{}lcccl@{}}
    \toprule
    \textbf{System / Model} & \textbf{Year} & \textbf{Update Dynamics} & \textbf{Persistence} & \textbf{Main Paradigm} \\ 
    \midrule
    \midrule
    \multicolumn{5}{c}{\textit{\textbf{Implicit Memory}}}\\
    \midrule
    Transformer (KV Cache)~\cite{vaswani2023attentionneed} & 2017 & Online & Short-Term & KV-based Attention \\
    Sparse Transformer~\cite{child2019sparse} & 2019 & Online & Short-Term & Sparse Attention Patterns \\
    Big Bird~\cite{zaheer2020bigbird} & 2020 & Online & Short-Term & Sparsified Attention \\
    Longformer~\cite{beltagy2020longformer} & 2020 & Online & Short-Term & Local and Global Attention \\
    StreamingLLM~\cite{xiao2023streamingllm} & 2023 & Online & Short-Term & Attention Sinks with Sliding Window \\
    RetNet~\cite{sun2023retentive} & 2023 & Online & Long-Term & Retentive Network (Recurrence) \\
    Mamba / Mamba-2~\cite{gu2024mamba} & 2024 & Online & Long-Term & State Space Recurrence \\
    RWKV (Eagle/Finch)~\cite{peng2024rwkv6} & 2024 & Online & Long-Term & Linear Recurrence with Matrix States \\
    Samba~\cite{ren2024samba} & 2024 & Online & Long-Term & Interleaved Mamba and Sliding Window \\
    Jamba / Jamba-1.5~\cite{lieber2024jambahybridtransformermambalanguage} & 2024 & Online & Long-Term & Hybrid SSM-Attention Blocks \\ 
    Selective Attention~\cite{leviathan2024selective} & 2024 & Online & Short-Term & Gated/Adaptive Attention \\
    LightTransfer~\cite{zhang2024light} & 2024 & Online & Short-Term & Layer-wise Hybrid Adaptation \\
    Expansion Span~\cite{nunez2025expansion} & 2025 & Online & Long-Term & Hybrid with Span-Expanded Attention \\
    MoBA~\cite{lu2025moba} & 2025 & Online & Short-Term & Content-Routed Sparse KV Blocks \\
    Gated Delta Networks~\cite{yang2025deltanet} & 2025 & Online & Long-Term & Gated Recurrence with Delta-Rule \\
    RWKV-7~\cite{peng2025rwkv7} & 2025 & Online & Long-Term & Vector-Gated Generalized Delta Recurrence \\
    Log-Linear Attention~\cite{guo2025loglinearattention} & 2025 & Online & Long-Term & Logarithmic Hierarchy of Recurrent States \\
    KDA / Kimi Linear~\cite{kimiteam2025kimilinearexpressiveefficient} & 2025 & Online & Long-Term & Channel-Wise Gated Delta Memory \\
    MoM~\cite{du2025momlinearsequencemodeling} & 2025 & Online & Long-Term & Routed Multi-State Recurrence \\
    Hybrid Quadratic-Linear Transformer~\cite{irie2025blendingcomplementarymemorysystems} & 2025 & Online & Long-Term & Fixed Complementary KV-RNN Memory \\
    RATTENTION~\cite{rattention2025} & 2025 & Online & Short-Term & Local-Global Minimal Sliding Window \\
    Mamba-3~\cite{lahoti2026mamba3} & 2026 & Online & Long-Term & Complex-Valued MIMO State Space Recurrence \\
    Gated DeltaNet-2~\cite{hatamizadeh2026gateddeltanet2decouplingerase} & 2026 & Online & Long-Term & Decoupled Delta-Rule State Editing \\
    Kaczmarz Linear Attention~\cite{zou2026kaczmarzlinearattention} & 2026 & Online & Long-Term & Projection-Calibrated Linear Attention \\
    Kalman Linear Attention / Gated KalmaNet~\cite{shaj2026kalmanlinearattentionparallel,peng2026gatedkalmanetfadingmemory} & 2026 & Online & Long-Term & Probabilistic Recurrent State Estimation \\
    Next-Latent Prediction~\cite{teoh2026nextlatentpredictiontransformerslearn} & 2026 & Offline & Long-Term & Latent Belief-State Induction \\
    HyperMLP~\cite{lu2026hypermlpintegratedperspectivesequence} & 2026 & Online & Short-Term & Dynamic Sequence-Mixing Attention \\
    Memory Caching~\cite{behrouz2026memorycachingrnnsgrowing} & 2026 & Online & Long-Term & Cached Recurrent State Snapshots \\
    AMOR~\cite{zheng2026thinkfastslowamor} & 2026 & Online & Long-Term & Entropy-Gated Attention Refinement \\
    HAM~\cite{lufkin2026hybridassociativememories} & 2026 & Online & Long-Term & Surprise-Gated KV-RNN Hybrid Memory \\
    \midrule
    \midrule
    \multicolumn{5}{c}{\textit{\textbf{Explicit Memory}}}\\
    \midrule
    kNN-LM~\cite{khandelwal2020nearest} & 2021 & Offline & Long-Term & Retrieval-Augmented \\
    Switch Transformer~\cite{fedus2021switch} & 2021 & Offline & Long-Term & Conditional Parameter (Sparse MoE) \\
    GLaM~\cite{du2022glam} & 2022 & Offline & Long-Term & Conditional Parameter (Sparse MoE) \\
    PlugLM~\cite{cheng2023decoupleknowledgeparameters} & 2023 & Offline & Long-Term & Editable Plug-in KV Memory \\
    Mixtral / Mixtral 8x7B~\cite{mixtral2024} & 2024 & Offline & Long-Term & Conditional Parameter (MoE) \\
    DeepSeek-MoE~\cite{deepseek2024moe} & 2024 & Offline & Long-Term & Conditional Parameter (MoE) \\
    MEMORYLLM~\cite{wang2024memoryllmselfupdatablelargelanguage} & 2024 & Online & Long-Term & Persistent Latent Memory Pool \\
    LM2~\cite{kang2025lm2largememorymodels} & 2025 & Offline & Long-Term & Dedicated Memory Slots \\
    ExplicitLM~\cite{yu2025explicitlmdecouplingknowledgeparameters} & 2025 & Offline & Long-Term & Interpretable Memory Banks \\
    TTT-E2E~\cite{tandon2025endtoend} & 2025 & Online & Long-Term & End-to-End Test-Time Training \\
    Titans~\cite{behrouz2025titans} & 2025 & Online & Long-Term & Parametric with Test-Time Training \\
    Nested Learning~\cite{behrouz2025nested} & 2025 & Online & Long-Term & Multi-Timescale Updated Modules \\
    Hydra~\cite{chaudhary2025hydramodulararchitectureefficient} & 2025 & Online & Long-Term & Multi-Component (SSM + MoE) \\
    Engram~\cite{cheng2026conditionalmemoryscalablelookup} & 2026 & Offline & Long-Term & Hashed Lookup Slots \\
    In-Place TTT~\cite{feng2026inplacetesttimetraining} & 2026 & Online & Long-Term & Fast-Weight MLP Update \\
    GDWM~\cite{mei2026gateddifferentiableworkingmemory} & 2026 & Online & Long-Term & Gated Test-Time Memory Update \\
    MemoryLLM (FFN)~\cite{jaiswal2026memoryllmplugnplayinterpretablefeedforward} & 2026 & Offline & Long-Term & Context-Free Feed-Forward Lookup \\
    \bottomrule
    \end{tabular}
}
\end{table*}

\subsection{Representation: Implicit vs.\ Explicit Memory}
\label{subsec:representation}

The first axis concerns how memory is represented and accessed within the model. We distinguish between \emph{implicit} and \emph{explicit} memory based on whether the memory is endowed with an independent and controllable interface for storage or retrieval.

\paragraph{Implicit Memory}
Implicit memory refers to information that is stored as a byproduct of the model's forward computation and is tightly coupled with its internal dynamics. Such memory does not expose an explicit read, write or lookup interface; instead, it is accessed implicitly through the computation graph. Typical examples include evolving states in recurrent sequence memory---as instantiated by RNNs, linear attention, and SSMs---and key--value (KV) caches in Transformer-based models. Although some implicit memory structures (e.g., KV caches) are materialized as explicit tensors, their access semantics are fixed by the model architecture and lack independent addressing or update control. As a result, they function as transient computational states rather than stand-alone storage modules.

\paragraph{Explicit Memory}
Explicit memory, in contrast, is instantiated as a distinct storage component with well-defined access semantics. It typically supports explicit addressing, retrieval, and update operations that are not reducible to standard forward computation or offline backpropagation. Explicit memory may take the form of external tables, associative buffers, or dedicated memory modules integrated into the model. Notably, explicit memory is not restricted to non-parametric structures. Some approaches realize explicit memory through additional parameterized modules that are trained or updated with specialized rules, often at inference time. What distinguishes them from standard model parameters is not their parametric form, but the fact that they are \emph{designed and operated as memory}, with clear read/write semantics and controllable update policies.

\paragraph{Scope Clarification}
In this survey, we do not discuss static model parameters learned during pretraining (e.g., the weights of attention or MLP layers), even though they encode knowledge. Our focus is on memory mechanisms that evolve over time, or play an active role in storing and retrieving contextual or experiential information beyond what is captured by fixed base parameters. This representation axis allows us to place a wide range of methods---from attention-based context accumulation to retrieval-augmented and inference-time adaptive models---within a unified conceptual space, without conflating architectural form with functional role.

\subsection{Update Dynamics: Offline vs.\ Online Memory}
\label{subsec:update}

The second axis characterizes when and how (frequent) the memory is updated. We mainly distinguish between \emph{offline} and \emph{online} memory according to whether updates occur exclusively during training or can also take place at inference time.

\paragraph{Offline Memory}
Offline memory systems are updated only during a distinct model training phase, typically via gradient-based optimization. Once training is complete, the memory remains fixed during inference. Many parameterized memory components, including certain retrieval embeddings or memory-augmented layers trained end-to-end, fall into this category.

\paragraph{Online Memory}
Online memory systems support updates during inference, allowing the model to incorporate new information on the fly. Such updates may be triggered at various frequencies: per token, per segment, or per batch, and can be governed by specialized rules distinct from standard backpropagation. Online memory enables adaptation to non-stationary inputs and long-horizon dependencies, and plays a central role in recent work on test-time training and inference-time plasticity.

\paragraph{Fine-Grained Update Rules}
The offline/online distinction specifies \emph{when} memory can be modified, but it does not fully characterize \emph{how} the memory content is updated. Two memory mechanisms may both be online while relying on very different update rules: a recurrent state may be revised by a closed-form state transition, a fast-weight module may be optimized by gradients, and a hybrid model may update its accessible KV set only when an uncertainty or error signal exceeds a threshold. We therefore treat update rules as a secondary lens within the update-dynamics axis, as shown in Table~\ref{tab:update_rules}. This refinement preserves the primary taxonomy while making the dynamics of memory writing, retention, and forgetting more explicit. It is worth noting that these update rules are not mutually exclusive. For instance, some memory architectures combine optimization-based writing with surprise-driven modulation, and some combine recurrent state-transition updates with signal-gated admission into a sparse KV cache. The purpose of this refinement is therefore not to introduce a fourth primary axis, but to expose the mechanisms that determine how memory changes once an update opportunity is available.

\begin{table*}[!t]
\centering
\caption{A fine-grained view of memory update rules. The offline/online axis describes when updates are available, while the update rule describes the mechanism by which memory content, access, or state is modified.}
\label{tab:update_rules}
\footnotesize
\setlength{\tabcolsep}{3pt}
\renewcommand{\arraystretch}{1.12}
\begin{tabular}{@{}>{\raggedright\arraybackslash}p{0.16\textwidth}
                >{\raggedright\arraybackslash}p{0.31\textwidth}
                >{\raggedright\arraybackslash}p{0.29\textwidth}
                >{\raggedright\arraybackslash}p{0.17\textwidth}@{}}
\toprule
\textbf{Update Rule} & \textbf{Mechanism} & \textbf{Representative Examples} & \textbf{Main Risks} \\
\midrule
Optimization-based writing & Memory parameters or fast weights are updated by minimizing an explicit objective, either during training or at inference time. & TTT-E2E~\cite{tandon2025endtoend}, Titans~\cite{behrouz2025titans}, In-Place TTT~\cite{feng2026inplacetesttimetraining}, Nested Learning~\cite{behrouz2025nested} & Drift, instability, and objective mismatch \\
\cdashline{1-4}[2pt/2pt]
State-transition updates & A hidden or recurrent state is updated by a learned transition, delta rule, projection, or filtering equation. & Mamba~\cite{gu2024mamba}, Gated Delta Networks~\cite{yang2025deltanet}, Gated DeltaNet-2~\cite{hatamizadeh2026gateddeltanet2decouplingerase}, Kaczmarz Linear Attention~\cite{zou2026kaczmarzlinearattention}, Kalman Linear Attention~\cite{shaj2026kalmanlinearattentionparallel} & Compression loss and state interference \\
\cdashline{1-4}[2pt/2pt]
Signal-gated writing or routing & A surprise, uncertainty, prediction-error, or utility signal controls whether a memory write, cache admission, or expensive memory path is activated. & Titans~\cite{behrouz2025titans}, GDWM~\cite{mei2026gateddifferentiableworkingmemory}, AMOR~\cite{zheng2026thinkfastslowamor}, HAM~\cite{lufkin2026hybridassociativememories} & Calibration, threshold sensitivity, and delayed relevance \\
\cdashline{1-4}[2pt/2pt]
Admission, eviction, and consolidation & The memory store is modified by deciding which entries are admitted, retained, compressed, evicted, or rewritten. & StreamingLLM~\cite{xiao2023streamingllm}, RATTENTION~\cite{rattention2025}, HAM~\cite{lufkin2026hybridassociativememories}, Bottlenecked Transformers~\cite{oomerjee2026bottleneckedtransformersperiodickv} & Irreversible loss and retrieval bias \\
\cdashline{1-4}[2pt/2pt]
Objective-induced or structural updates & Memory behavior is shaped by auxiliary objectives, architectural schedules, layer replacement, or offline construction rather than by per-instance online writes. & Engram~\cite{cheng2026conditionalmemoryscalablelookup}, MoE models~\cite{fedus2021switch,deepseek2024moe}, Priming~\cite{chattopadhyay2026priminghybridstatespace}, Next-Latent Prediction~\cite{teoh2026nextlatentpredictiontransformerslearn} & Rigidity, objective mismatch, and limited test-time adaptability \\
\bottomrule
\end{tabular}
\end{table*}

\subsection{Persistence: Short-Term vs.\ Long-Term Memory}
\label{subsec:persistence}

The third axis concerns the effective temporal horizon over which stored information can continue to influence computation. This horizon is determined not only by whether a memory state is physically retained, but also by its storage granularity and compression rate. A memory mechanism may store many high-fidelity local items but remain short-horizon because its storage grows quickly with sequence length; conversely, a compact recurrent or parameterized state may have limited physical size while maintaining a much longer contextual view through aggressive compression.

\paragraph{Short-Term Memory}
Short-term or transient memory stores information at relatively fine granularity and is typically bounded to a local context window, segment, or inference episode. Attention KV caches are the canonical example: they preserve token-level states with high fidelity and low compression, but their storage grows with the number of retained tokens and therefore becomes difficult to scale indefinitely. Such mechanisms can support precise local recall, yet their effective horizon is limited by window size, cache budget, eviction policy, or session boundaries.

\paragraph{Long-Term Memory}
Long-term memory is designed to preserve information or its influence over extended contexts, and in some cases across sessions. It does not necessarily require a large uncompressed store. Recurrent sequence memory, for example, may maintain only a fixed-size state, but this state acts as a compressed summary whose receptive field can span very long sequences. Explicit stores, memory slots, and test-time updated parameters provide another route by retaining addressable or adaptable information beyond a local window. Thus, long-term memory is characterized by an extended effective horizon, often enabled by higher compression, selective retention, or persistent storage, rather than by storage size alone.

\paragraph{Interplay of Axes}
These three axes are largely orthogonal. For instance, explicit memory can be either offline or online, and online memory may exhibit either short-term or long-term persistence depending on its update and eviction policies. By decoupling representation, update dynamics, and persistence, our taxonomy highlights the design space of memory mechanisms in large language models and provides a structured lens for analyzing existing and future approaches.

\subsection{Positioning Relative to Memory-Centered Surveys}
\label{subsec:survey_positioning}

As illustrated in Table~\ref{tab:survey_comparison}, recent surveys use the term memory at different system levels, so their taxonomies are complementary rather than directly interchangeable. Early work centered on LLM-based agents organizes memory modules by their sources, forms, and operations, with particular attention to agent--environment interaction, applications, and evaluation~\cite{zhang2024agentmemorysurvey}. Cognitively inspired surveys instead relate human memory categories to AI systems. For example, Wu et al.~\cite{wu2025humanmemorysurvey} classify memory along object (personal/system), form (parametric/non-parametric), and time (short/long term). Other work foregrounds the memory lifecycle: Du et al.~\cite{du2025rethinkingmemory} combine representation types with atomic operations such as consolidation, updating, indexing, forgetting, retrieval, and compression. A broader LLM-level treatment by Zhang et al.~\cite{zhang2025llmmemorysurvey} spans parametric, contextual, external, and procedural/episodic memory, emphasizing evaluation protocols and governance. More recently, Luo et al.~\cite{luo2026storageexperience} characterize the evolution of agent memory from trajectory storage, through reflection, to reusable experience.

\begin{table*}[t]
\centering
\caption{Positioning of this survey relative to representative memory-centered surveys. The comparison records each work's primary scope and organizing lens rather than every topic it covers.}
\label{tab:survey_comparison}
\footnotesize
\setlength{\tabcolsep}{4pt}
\renewcommand{\arraystretch}{1.12}
\begin{tabular}{@{}>{\raggedright\arraybackslash}p{0.13\textwidth}
                    >{\raggedright\arraybackslash}p{0.05\textwidth}
                    >{\raggedright\arraybackslash}p{0.35\textwidth}
                    >{\raggedright\arraybackslash}p{0.41\textwidth}@{}}
\toprule
\textbf{Survey} & \textbf{Year} & \textbf{Scope and organizing lens} & \textbf{Main emphasis / distinction} \\
\midrule
Zhang et al.~\cite{zhang2024agentmemorysurvey} & 2024 & LLM-based agents; memory sources, forms, and operations. & Design and evaluation of agent memory modules for long-horizon interaction and downstream applications. \\
\cdashline{1-4}[2pt/2pt]
Wu et al.~\cite{wu2025humanmemorysurvey} & 2025 & LLM-driven AI systems; object, form, and time (3D--8Q). & Correspondence between human-memory categories and personal/system, parametric/non-parametric, and short/long-term AI memory. \\
\cdashline{1-4}[2pt/2pt]
Du et al.~\cite{du2025rethinkingmemory} & 2025 & LLM-based AI and agents; representation types and six atomic operations. & Lifecycle-level management, including consolidation, indexing, updating, forgetting, retrieval, and compression. \\
\cdashline{1-4}[2pt/2pt]
Zhang et al.~\cite{zhang2025llmmemorysurvey} & 2025 & LLM memory across the model lifecycle; parametric, contextual, external, and procedural/episodic memory. & Operational definitions, evaluation protocols, editing/forgetting, governance, and auditing. \\
\cdashline{1-4}[2pt/2pt]
Luo et al.~\cite{luo2026storageexperience} & 2026 & LLM-agent memory; Storage--Reflection--Experience evolution. & Transformation of interaction trajectories into refined and transferable agent experience. \\
\hline
\blue{\textbf{This survey}} & \blue{2026} & \blue{\textbf{Model-level LLM architectures; representation, update dynamics, and persistence.}} & \blue{\textbf{How architectural substrates---attention/KV states, recurrent sequence states, writable parameters, addressable modules, and hybrids---store, update, and retain information during model computation.}} \\
\bottomrule
\end{tabular}
\end{table*}

Our focus is therefore deliberately centered on model-level architectural mechanisms, complementing rather than superseding system- and cognition-centered accounts. Conversational records, user profiles, reflection traces, tool outputs, and database-backed stores are important forms of memory in broader LLM and agent systems. They enter the scope of this survey, however, only when their storage, access, or update mechanisms are instantiated within, or tightly coupled to, the model architecture. Similarly, semantic categories such as factual and episodic memory provide a useful functional perspective, whereas our primary question is how information is architecturally represented, whether its update is tied to forward dynamics or governed by a distinct write process, and what effective temporal horizon it can sustain. This lens places mechanisms that are often separated across long-context, efficient-sequence-modeling, test-time-learning, and external-memory literatures into a common design space. In particular, it permits direct comparison between attention caches, recurrent sequence compression, inference-time writable modules, and architecturally integrated lookup stores while maintaining a clear interface with agent-level memory management. The contribution is thus not a replacement for agent- or cognition-centered taxonomies, but a complementary, mechanism-centered account of memory as an architectural property of the LLM itself.

\section{Implicit Memory via Computation Dynamics}
\label{sec:implicit}

Implicit memory in large language models arises from the internal dynamics of computation. Rather than being stored in independently persistent parameters or external addressable repositories, it is instantiated through transient activations, attention interactions, and evolving hidden states during forward execution. Such memory is tightly coupled to the computational graph and typically discarded once inference terminates. In this section, we examine implicit memory mechanisms through three complementary lenses. First, we analyze attention-based representations that encode prior context within token-to-token interactions. Second, we discuss runtime memory management strategies that govern how such contextual information is retained or pruned under long-sequence and streaming settings. Third, we review recurrent sequence memory, which compresses history into structured dynamical states. We conclude by examining structural limitations shared by implicit memory systems.

\subsection{Attention as Implicit Memory}
\label{subsec:attention_memory}

Attention mechanisms provide the canonical instantiation of implicit memory in modern transformer-based large language models~\cite{vaswani2023attentionneed}. Rather than maintaining a separate storage module, self-attention realizes memory through the forward computation itself: representations of past tokens are calculated and retained as key--value pairs, and incoming queries retrieve relevant information via content-based interaction.

\paragraph{Attention as Content-Addressable Transient Memory}
At a high level, self-attention can be interpreted as a differentiable, content-addressable transient working memory. During autoregressive inference, models maintain a key--value (KV) cache that accumulates representations of previously processed tokens. Each new token implicitly updates this memory by appending new entries, while retrieval is performed through attention weights computed on the fly. Although KV caches are explicitly materialized as tensors, they lack independent addressing or controllable read/write semantics; their evolution is fully determined by the internal forward pass. Consequently, they function as implicit computational memory rather than explicit memory modules.

\begin{figure*}[t!]
    \centering
    \includegraphics[width=0.95\linewidth]{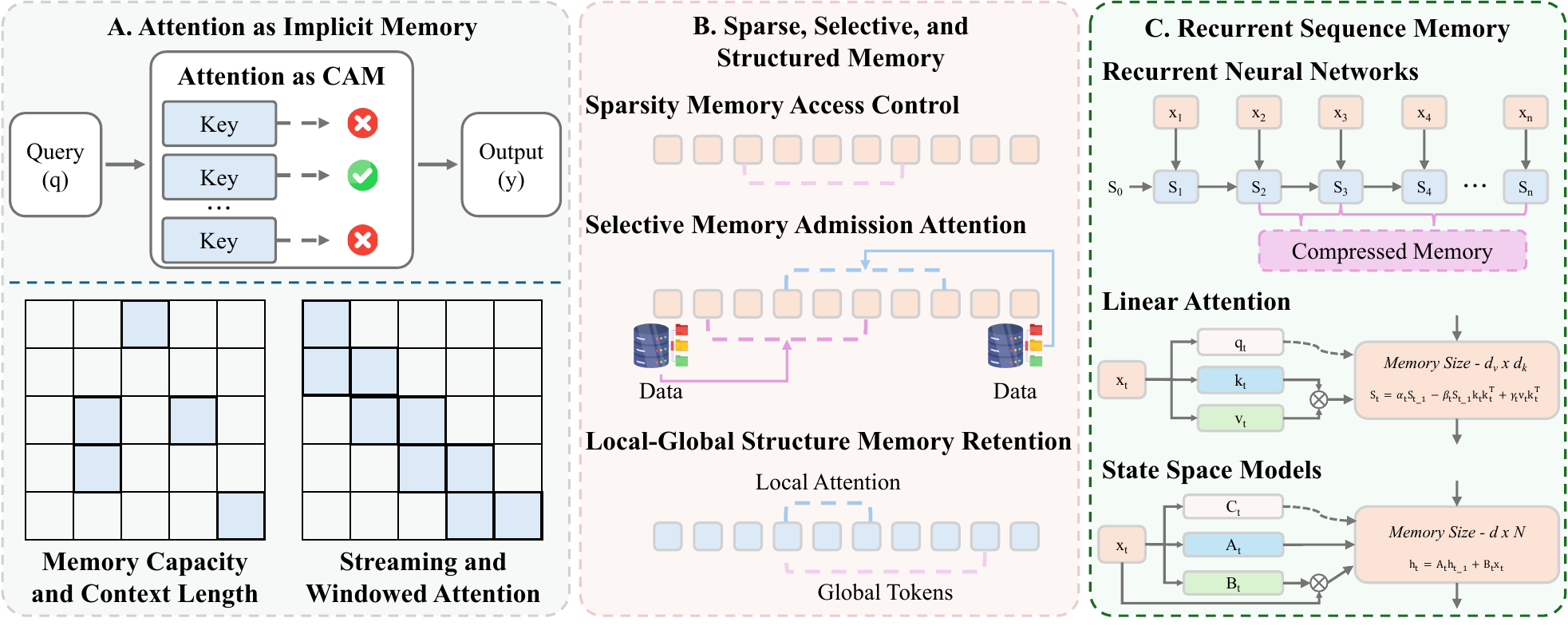}
    \caption{Overview of implicit memory via computation dynamics.}
    \label{fig:implicit_overview}
\end{figure*}

\paragraph{Memory Capacity and Context Length}
The capacity of attention-based implicit memory is directly tied to the model's context window. Increasing the context length effectively enlarges the short-term memory available to the model, enabling it to retain and utilize information over longer sequences. Some approaches reduce the per-token storage footprint: Multi-head Latent Attention (MLA) in DeepSeek-V2~\cite{deepseekai2024deepseekv2} jointly compresses keys and values into latent vectors, changing the granularity and efficiency---but not the implicit semantics---of KV memory. Sparse attention architectures instead reduce computation by restricting access to structured subsets of the context~\cite{child2019sparse,zaheer2020bigbird,beltagy2020longformer}. Recent studies further demonstrate that extending context length does not uniformly translate into effective memory utilization. Long-context models often fail to exploit their nominal capacity, revealing a gap between theoretical memory size and practical usage~\cite{hsieh2024rulerwhatsrealcontext,zhang2024inftybenchextendinglongcontext}. Attention-based implicit memory is therefore constrained not only by architectural resources, but also by optimization and inductive biases.

\paragraph{Streaming and Windowed Attention}
Beyond static context extension, streaming attention methods aim to support inference over arbitrarily long sequences with bounded memory. StreamingLLM introduces a simple yet effective strategy that combines sliding-window attention with a small set of preserved attention sink tokens, enabling stable generation over unbounded input streams without retraining \cite{xiao2023streamingllm}. From the viewpoint of implicit memory, such methods implement implicit eviction policies: most past states are continuously overwritten, while a limited subset is retained to anchor attention and prevent degradation. Related approaches, including ring-based and local--global attention schemes, similarly manage implicit memory by constraining which past tokens remain accessible at each step \cite{rattention2025}. Despite differences in implementation, these methods share a common principle: long-horizon behavior emerges from carefully designed access patterns over a fundamentally short-term, online memory.

\paragraph{Alternative Views of Attention as Memory Computation}
Recent work also revisits attention itself as a more general memory computation primitive rather than a normalized retrieval rule. HyperMLP~\cite{lu2026hypermlpintegratedperspectivesequence} reformulates autoregressive attention as an input-conditioned two-layer MLP whose effective hidden width grows with sequence length, replacing the conventional probability-simplex interpretation with dynamic sequence mixing. From the perspective of our taxonomy, HyperMLP still belongs to implicit memory: historical information is stored and accessed through the forward computation itself rather than through a separately addressable memory module. Its significance lies less in changing the memory category than in showing that even within attention-style implicit memory, the geometry of memory access and sequence mixing can be substantially generalized.

\paragraph{Summary and Limitations}
Attention-based mechanisms exemplify the strengths of implicit memory: seamless integration with model computation, online updates, and flexible content-based access. At the same time, their memory is inherently short-term, tightly bounded by architectural design, and difficult to control or persist across extended horizons. 

\subsection{Sparse, Selective, and Structured Memory}
\label{subsec:sparse_memory}

While standard self-attention treats all past tokens as equally admissible memory entries, a substantial body of work explores mechanisms that impose structure, sparsity, or selectivity on attentional memory. From the perspective of implicit memory, these approaches can be interpreted as introducing \emph{admission}, \emph{access}, and \emph{retention} control over an online memory system.

\paragraph{Sparsity as Memory Access Control}
Early sparse attention architectures demonstrate that full attention over all past tokens is often unnecessary. Methods such as Sparse Transformers~\cite{child2019sparse}, BigBird~\cite{zaheer2020bigbird}, and Longformer~\cite{beltagy2020longformer} restrict attention to structured subsets of the sequence---including local windows, strided patterns, or a small number of global tokens---to reduce computational cost while preserving performance on long sequences. Viewed through the lens of memory, these sparsity patterns define which memory entries are accessible at each step. Rather than increasing memory capacity, sparse attention reshapes access paths within a fixed implicit memory budget, prioritizing certain regions of the context while ignoring others. Importantly, memory updates remain fully online and implicit: all tokens are still encoded, but only a subset can be retrieved at any given time.

\paragraph{Content-Routed Sparse Attention}
Recent sparse mechanisms replace fixed access patterns with learned routing over coarse memory regions. MoBA~\cite{lu2025moba} partitions the KV context into blocks and routes each query to a small set of relevant blocks, translating MoE-style conditional computation into block-level memory access. Native Sparse Attention (NSA)~\cite{yuan2025nativesparseattention} combines compressed block summaries, selectively retained fine-grained blocks, and a local window in a hardware-aligned design. Both remain attention-based implicit memory: they preserve token-derived KV representations but reduce the portion read by each query. Their primary contribution is therefore adaptive access control rather than a new persistent storage substrate.

\paragraph{Selective Attention and Memory Admission}
Beyond static sparsity patterns, more recent work explores learned or adaptive mechanisms that determine which tokens should be emphasized or retained. Selective attention models introduce data-dependent gating or routing functions that modulate attention scores, effectively deciding which contextual information should influence the current computation \cite{leviathan2024selective}. From a memory standpoint, such mechanisms act as implicit admission controls: although representations of all tokens exist within the computation graph, only selected entries are amplified and meaningfully contribute to downstream processing. This selectivity allows the model to focus its limited short-term memory capacity on salient information without introducing an explicit memory store or retrieval interface.

\paragraph{Local--Global Structure and Memory Retention}
Another prominent line of work organizes attention into local and global components to balance short-range detail and long-range dependency modeling. Local attention mechanisms emphasize recent tokens, while global tokens or summaries provide a coarse but persistent view of the broader context. Architectures such as BigBird and related local--global attention schemes exemplify this design \cite{zaheer2020bigbird}. More recent studies further investigate how minimal global memory is required to stabilize long-context inference. For example, RATTENTION systematically analyzes the trade-off between sliding window size and the number of retained global tokens, showing that a small but carefully chosen global memory can substantially improve performance \cite{rattention2025}. These findings highlight that long-horizon behavior in implicit memory systems often arises not from unbounded storage, but from structured retention of a limited set of memory anchors.

\paragraph{Unifying View}
Across sparsity, selectivity, and structural decomposition, a common theme emerges: due to storage or computational constraints, attentional memory mechanisms often do not scale by simply storing more information, but by carefully controlling how information is admitted, accessed, and retained within the forward computation. These controls remain implicit---they are embedded in attention patterns and internal dynamics rather than exposed as explicit memory operations. As a result, such approaches preserve the efficiency and simplicity of implicit memory while partially mitigating its capacity limitations.

\subsection{Recurrent Sequence Memory}
\label{subsec:ssm_memory}

Beyond attention, recurrent sequence memory compresses history into an evolving latent state coupled to forward computation. It is instantiated by classical recurrent neural networks (RNNs), linear-attention recurrences, and state space models (SSMs). Classical RNNs typically learn relatively unconstrained nonlinear state transitions, whereas SSMs derive recurrences from structured dynamical systems---often linear in the state with learned or input-dependent modulation---that facilitate efficient parallel scans and controlled long-range dynamics. Modern linear recurrent models increasingly blur this boundary, but the distinction remains useful for understanding their architectural inductive biases.

\paragraph{Recurrent Neural Networks}
Classical recurrent neural networks (RNNs) are among the earliest forms of implicit memory, compressing prior inputs into a hidden state updated at each timestep~\cite{elman1990finding}. Long short-term memory (LSTM) networks extend its effective horizon through gated information flow~\cite{hochreiter1997lstm}. In both cases, memory is stored and accessed entirely through state transitions.

\paragraph{Linear Attention and Structured State Models}
Linear Attention established an important bridge from attention to recurrent memory: by kernelizing attention and exploiting associativity, it expresses causal attention as a recurrent state updated in linear time~\cite{pmlr-v119-katharopoulos20a}. Gated Linear Attention (GLA) adds data-dependent decay to this matrix-valued state~\cite{pmlr-v235-yang24ab}. RWKV likewise combines parallel training with recurrent inference; RWKV-5/6 introduce matrix-valued states and dynamic recurrence~\cite{peng2024rwkv6}. State space models (SSMs) generalize this state-based view through structured dynamical systems. Mamba and Mamba-2 introduce selective updates and efficient State Space Duality formulations~\cite{gu2024mamba,dao2024mamba2}, trading flexible token-wise access for compressed temporal dynamics. Mamba-3~\cite{lahoti2026mamba3} enriches the state through improved discretization, complex-valued updates, and a multi-input multi-output formulation, while preserving online fixed-size storage. Gated Delta Networks (GDNs)~\cite{yang2025deltanet} instead use input-conditioned delta-rule updates for content-aware retention.

\paragraph{State Editing, Projection, and Filtering}
Recent work gives recurrent updates more structured editing semantics. Gated DeltaNet-2~\cite{hatamizadeh2026gateddeltanet2decouplingerase} separates key-side erasure from value-side writing, while Kaczmarz Linear Attention~\cite{zou2026kaczmarzlinearattention} casts writes as key-normalized projection steps. Kalman Linear Attention~\cite{shaj2026kalmanlinearattentionparallel} and Gated KalmaNet~\cite{peng2026gatedkalmanetfadingmemory} instead use filtering or online regression to modulate updates by estimation quality. All remain implicit because the state and its update rule are coupled to forward dynamics.

\paragraph{Expressive Delta and Bilinear Updates}
Several models increase recurrent-memory expressivity by refining how a token transforms a matrix-valued state. RWKV-7~\cite{peng2025rwkv7} generalizes the delta rule with vector-valued state gates and in-context learning rates while decoupling removal and addition keys, enabling channel-wise state replacement. Kimi Delta Attention (KDA), the recurrent core of Kimi Linear~\cite{kimiteam2025kimilinearexpressiveefficient}, instead introduces channel-wise diagonal decay within a structured diagonal-plus-low-rank transition. DeltaProduct~\cite{siems2025deltaproduct} composes multiple Householder updates to improve state tracking, whereas Comba~\cite{hu2025comba} uses scalar-plus-low-rank transitions and closed-loop feedback. These variants share a common direction: replacing uniform decay or rank-one correction with richer, input-dependent state editing, while retaining bounded recurrent memory.

\paragraph{Hierarchical State Capacity}
Other work changes how recurrent capacity is allocated. Log-Linear Attention~\cite{guo2025loglinearattention} replaces one fixed-size state with a Fenwick-tree hierarchy that grows logarithmically with sequence length, preserving recent information at finer resolution and distant history in coarser summaries. It therefore occupies a middle ground between constant-state linear attention and token-wise KV storage.

\paragraph{Hybrid and Recurrent--Attention Variants}
Hybrid designs combine recurrent compression with more flexible computation. RetNet offers recurrent and chunkwise forms~\cite{sun2023retentive}, while Jamba and Jamba-1.5 interleave attention with Mamba blocks~\cite{lieber2024jambahybridtransformermambalanguage,jambateam2024jamba15hybridtransformermambamodels}. MoM~\cite{du2025momlinearsequencemodeling} instead routes tokens across multiple recurrent states to reduce over-compression and interference. These designs balance efficient state propagation with selective access or allocation.

\paragraph{Training-Induced Belief States}
Training objectives can also shape implicit memory without changing its storage substrate. Next-Latent Prediction~\cite{teoh2026nextlatentpredictiontransformerslearn} adds an objective for predicting future latent states, encouraging hidden representations to preserve history useful for future prediction. It is therefore an offline, objective-induced mechanism rather than a new test-time write rule.

\paragraph{Update-Rule Perspective}
Viewed through the refined update taxonomy in Table~\ref{tab:update_rules}, most recurrent sequence memory mechanisms instantiate \emph{state-transition updates}: memory is modified by a learned recurrence, selective scan, delta-rule correction, projection step, or filtering equation. The main difference is how each token changes and allocates the compressed state. Mamba-style models emphasize structured state transitions; RWKV-7, KDA, and related delta variants enrich erasure and writing; Log-Linear Attention distributes updates across hierarchical states; and Kalman-style variants introduce uncertainty-aware filtering. Next-Latent Prediction occupies a different position: it is primarily an offline objective-induced mechanism that changes the training pressure on hidden states rather than adding a new inference-time write rule.

\paragraph{Implications}
Recurrent sequence memory supports online processing of unbounded streams with bounded storage. However, its computation-coupled states lack flexible read/write control, and long-horizon compression can reduce the fidelity of retained information.

\subsection{Limitations of Implicit Memory}
\label{subsec:implicit_limitations}

Despite their effectiveness and efficiency, implicit memory mechanisms share fundamental limitations that stem from their tight coupling with model computation. These limitations are largely architectural in nature and manifest consistently across attention-based and state-based designs.

\paragraph{Bounded and Hard-to-Control Capacity}
Implicit memory is inherently constrained by architectural resources. In attention-based models, memory capacity scales with the context window and KV cache size, which are bounded by quadratic or linear-time complexity and hardware limits \cite{child2019sparse, zaheer2020bigbird, beltagy2020longformer}. Even when long-context architectures are employed, empirical studies show that models often fail to effectively utilize their nominal context length, revealing a gap between theoretical capacity and practical memory usage \cite{hsieh2024rulerwhatsrealcontext, zhang2024inftybenchextendinglongcontext}. State-based models, while capable of unbounded sequence processing in principle, compress historical information into fixed-size latent states. This compression introduces an inherent trade-off between memory span and representational fidelity, making it difficult to selectively preserve fine-grained past information over long horizons \cite{gu2024mamba, dao2024mamba2}.

\paragraph{Lack of Adaptable Read/Write Semantics}
A defining characteristic of implicit memory is the absence of explicit, adaptable, and controllable memory operations. Information is stored and retrieved only as dictated by the computation graph, whether through attention weights or state transitions. As a result, implicit memory may be insufficiently adaptive to input.

\paragraph{Limited Persistence Across Contexts}
Contents of implicit memory are often tied to a few forward passes or inference sessions and are typically discarded once the computation ends. Even mechanisms that enlarge effective recurrent capacity, such as cached state snapshots or mixtures of recurrent states, primarily improve retention within an extended sequence rather than introducing fully persistent memory across sessions \cite{behrouz2026memorycachingrnnsgrowing,du2025momlinearsequencemodeling}. While architectural modifications can extend the effective horizon of implicit memory, persistence across documents, tasks, or interactions remains difficult to achieve without introducing additional storage mechanisms. This limitation is particularly salient in settings that require continual adaptation or accumulation of experience beyond a session.

\paragraph{Implications for Memory Design}
These limitations do not diminish the importance of implicit memory; rather, they delineate the boundaries of what can be achieved through computation-coupled memory alone. The challenges of capacity control, persistence, and selective access motivate the development of memory mechanisms that decouple storage from internal computation and expose broader explicit semantics. Such mechanisms form the basis of explicit memory systems, which we examine in the next section.

\begin{figure*}[t!]
    \centering
    \includegraphics[width=0.9\linewidth]{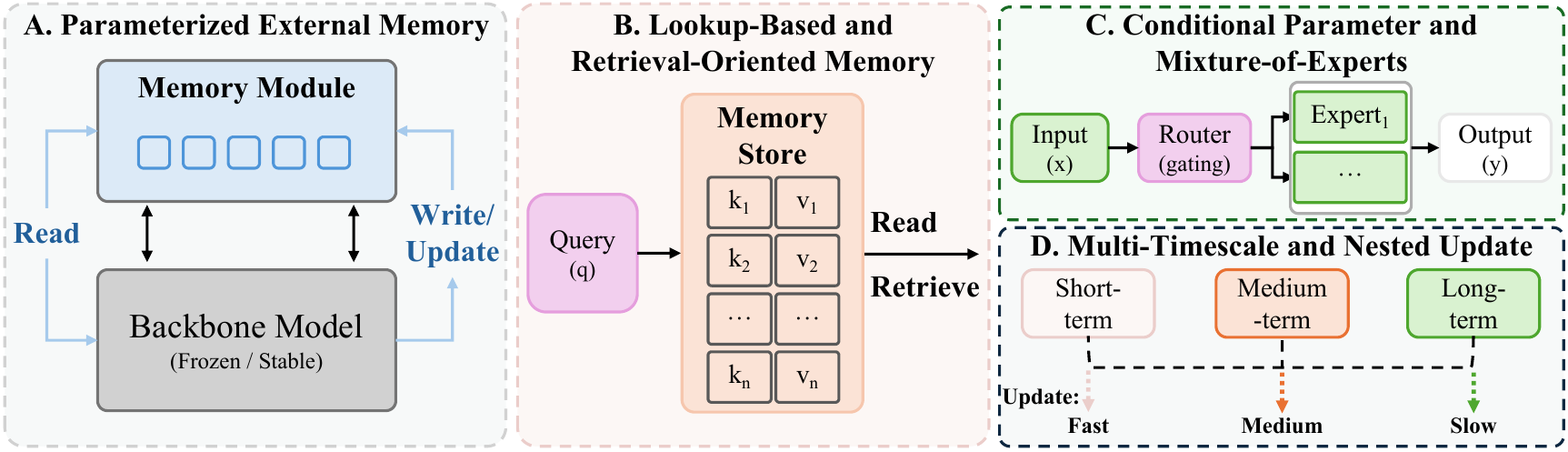}
    \caption{Overview of explicit memory via addressable and adaptive storage.}
    \label{fig:explicit_overview}
\end{figure*}
\section{Explicit Memory via Addressable and Adaptive Storage}
\label{sec:explicit}

While implicit memory is basically embedded within computation dynamics and discarded once inference session terminates, explicit memory introduces storage structures that are decoupled from transient forward states and often support more controllable access or persistence. In this paradigm, memory is not merely a by-product of activations or hidden-state evolution, but is instantiated as parameterized modules, structured key–value repositories, or dedicated storage components whose contents may persist across inference instances. Crucially, the defining property of explicit memory is not its implementation form or lifespan alone, but its autonomy: the stored information exists independently of the immediate computational graph and can be updated, accessed, or maintained under mechanisms distinct from standard forward propagation.

In this part, we focus strictly on \emph{model-level} explicit memory mechanisms. We do not consider prompt-level memory engineering, retrieval-augmented generation pipelines, or external agent frameworks that rely on databases or orchestration systems. Instead, we examine architectures in which explicit memory is incorporated as a structural component of the model itself, typically through additional parameters, addressable slots, or learnable update rules that operate alongside—but are not reducible to—static base model weights. We organize explicit memory systems along four complementary perspectives. First, we review parameterized memory modules that introduce dedicated memory parameters updated at distinct temporal scales. Second, we examine lookup-based and retrieval-oriented mechanisms that maintain persistent, addressable storage structures. Third, we discuss conditional parameter memory systems, represented by the Mixture-of-Experts (MoE) paradigm, which dynamically activates a sparse subset of expert modules based on specific routing functions. Finally, we analyze multi-timescale and nested update strategies that regulate how memory evolves across inference and training. Collectively, these approaches aim to extend the temporal horizon of language models beyond computation-bound implicit memory, while introducing new challenges in stability, scalability, and interference.

\subsection{Parameterized External Memory Modules}
\label{subsec:parameterized}
A central instantiation of explicit memory introduces \emph{dedicated memory parameters} that are structurally distinct from the base model weights and governed by specific update dynamics. Unlike implicit memory, parameterized external memory persists across inference instances and can be selectively updated without modifying the full set of model parameters. This decoupling enables controlled online plasticity while preserving the stability of the pretrained backbone.

\paragraph{Fast Weights and Meta-Learned Memory}
The idea of augmenting neural networks with rapidly changing memory parameters traces back to early fast-weight formulations~\cite{schmidhuber1992learning,ba2016using}. In these models, a set of fast weights is updated on short timescales to store recent information, while slow weights encode long-term knowledge. Memory-Augmented Neural Networks (MANNs)~\cite{santoro2016meta} and Differentiable Neural Computers (DNCs)~\cite{graves2016hybrid} further developed this paradigm by introducing explicit memory matrices with learnable read/write operations. Although originally proposed in smaller-scale settings, these works establish a foundational principle: memory can be parameterized as an independent component whose evolution is distinct from standard gradient-based training.

\paragraph{Test-Time Parameter Updates}
Recent large language model research revisits this principle at scale. Titans~\cite{behrouz2025titans} introduces a dedicated memory module updated during inference via surprise-driven gradient signals. Importantly, these updates modify only the memory parameters rather than the full backbone, thereby preserving pretrained knowledge while enabling task-specific adaptation. From our taxonomy perspective, Titans exemplifies explicit memory because the stored information resides in persistent parameters that survive beyond a single forward computation and are governed by update rules distinct from standard pretraining. Similarly, Test-Time Training (TTT) and its end-to-end variants (e.g., TTT-E2E~\cite{tandon2025endtoend}) incorporate auxiliary parameter subsets that are optimized during inference on incoming data. These approaches treat memory as a small, adaptable parameter space that accumulates information across inputs. MEMORYLLM~\cite{wang2024memoryllmselfupdatablelargelanguage} offers a closely related but architecturally distinct design by introducing a fixed-size latent memory pool embedded across Transformer layers and refreshed through forward-only self-updates when new knowledge is read. Unlike gradient-based test-time adaptation, such updates directly rewrite dedicated memory tokens while keeping the backbone unchanged, further illustrating that explicit online memory can be realized through persistent latent slots rather than only through parameter optimization. Nested Learning~\cite{behrouz2025nested} further extends this paradigm by introducing multi-frequency update schedules, where different parameter subsets evolve at distinct temporal scales. Across these methods, the defining feature remains consistent: memory is instantiated as one or a few parameterized modules whose updates are temporally and functionally separable from the static base model weights.

\paragraph{Fast-Weight Reuse and Selective Writing}
Recent work also explores how online memory can be introduced into existing Transformer blocks without adding a completely separate memory module. In-Place Test-Time Training~\cite{feng2026inplacetesttimetraining} treats the down-projection matrix in the feed-forward layer as a reusable fast-weight memory, updating it over input chunks while keeping the base model otherwise fixed. This design is explicit in the functional sense: a designated parameter subset is operated as writable memory under a specialized test-time objective, even though it is implemented inside a standard MLP block. Its main implication for the taxonomy is that explicit memory need not always appear as a visually separate bank or table; it can also arise when a normally static parameter matrix is assigned autonomous write semantics during inference. Gated Differentiable Working Memory~\cite{mei2026gateddifferentiableworkingmemory} adds another layer of control by gating test-time updates according to contextual utility, so that memory writes are concentrated on chunks that are likely to improve long-context prediction. Such write controllers are important because the main difficulty of online explicit memory is not only how to store information, but also how to decide which observations are worth committing to a persistent or semi-persistent state. At the same time, the boundary between explicit fast-weight memory and implicit sequence dynamics should be drawn carefully. Recent analysis of TTT with key--value binding argues that a broad class of such inner-loop updates can be reformulated as learned linear attention~\cite{liu2026testtimetrainingkvbinding}. This does not invalidate the memory interpretation of TTT-style modules, but it suggests that only variants with a clearly designated, autonomously updated storage substrate should be treated as explicit memory in the main taxonomy.

\paragraph{Dedicated Memory Slots}
Another branch makes the memory component structurally explicit. LM2~\cite{kang2025lm2largememorymodels} augments Transformer layers with dedicated memory slots accessed through cross-attention-like interactions and regulated by gating mechanisms. Unlike pure retrieval systems, these slots are trained as part of the model and interact with hidden states throughout computation; unlike implicit recurrent states, they are architecturally designated as memory variables with separate storage roles. This places LM2 close to parameterized external memory modules in our taxonomy, while also illustrating a broader design trend: explicit memory can be integrated at layer level as a persistent representational workspace rather than attached only as an external datastore or post-hoc retrieval component.

\paragraph{Update-Rule Perspective}
Parameterized explicit memory highlights the distinction between update timing and update mechanism. Titans, TTT-E2E, and In-Place TTT are all online in the broad taxonomy, but their memory changes through optimization-based writing: a designated parameter subset is modified according to an auxiliary or task-aligned objective. GDWM adds signal-gated control on top of this pattern by deciding which chunks deserve gradient-based writes. MEMORYLLM and LM2 instead illustrate slot-oriented or forward-update variants, where the memory substrate is structurally separated from the backbone even when updates are not ordinary full-model fine-tuning. This separation is important because it makes explicit memory less a single mechanism than a family of write rules over autonomous storage components.

\paragraph{Decoupled Learning Dynamics}
A unifying property of parameterized external memory systems is the separation of \emph{contextual storage parameters} and \emph{general knowledge parameters}. The backbone model encodes general linguistic competence, while memory parameters capture contextual or task-specific information. This separation mitigates catastrophic interference, as updating memory does not overwrite pretrained representations. However, it introduces new design tradeoffs, including memory capacity limits, update stability, and the risk of overfitting to transient signals.

\paragraph{Relation to Implicit Memory}
It is worth emphasizing that parameterized memory modules may still participate in forward computation at every token. What distinguishes them from implicit memory is not usage frequency, but persistence and autonomy. Their contents are not ephemeral activations tied to a single inference trajectory; instead, they constitute a persistent, learnable online state space that can accumulate information across episodes. Taken together, parameterized external memory modules represent a principled approach to extending temporal adaptability in large language models. By separating persistent memory storage from the main parameter body, they provide a controllable mechanism for balancing stability and plasticity—an objective that cannot be achieved through computation-bound implicit memory alone.

\subsection{Lookup-Based and Retrieval-Oriented Memory}
\label{subsec:lookup}
Beyond parameterized memory modules, another class of explicit memory mechanisms introduces \emph{long-term, addressable storage structures} that are accessed via content-based retrieval. Unlike implicit attention, lookup-based memory maintains independently stored entries whose contents are not recomputed from scratch and can persist across inference episodes. Many such stores are constructed offline, but some architectures also support online admission of newly observed representations without updating the backbone weights.


\paragraph{Nearest-Neighbor and Datastore-Augmented Models}
In the context of large language modeling, kNN-LM~\cite{khandelwal2020nearest} demonstrates that augmenting a pretrained model with a persistent datastore of context--target pairs can substantially improve perplexity and rare-token prediction. Crucially, the datastore is not part of the standard forward computation graph: it is an externalized, addressable memory that survives across inference calls. Although its keys are neural context representations, its values correspond to observed next tokens from a corpus, and the datastore is constructed as an independently maintained non-parametric resource. From our taxonomy perspective, kNN-LM therefore exemplifies explicit memory more clearly than ordinary KV cache, because the knowledge it contributes resides in a stable repository queried through a separate nearest-neighbor mechanism.

\paragraph{Boundary with Activation-Level Retrieval}
Some retrieval-augmented Transformer variants occupy a more ambiguous position. Memorizing Transformers~\cite{wu2022memorizingtransformers} store internal key--value representations from past inputs in a non-differentiable memory and retrieve them with approximate nearest-neighbor search, while LongMem~\cite{wang2023augmentinglanguagemodelslongterm} stores key--value pairs extracted from a frozen backbone into a cache memory bank and uses a SideNet to retrieve and fuse them with the current context. These designs introduce explicit indexing and retrieval machinery, but the stored content is still primarily a materialized intermediate state of the model computation. In this respect, they are better viewed as a boundary class between explicit datastore memory and implicit activation memory: they extend the effective reach of KV-style context storage, yet do not expose the same degree of semantic autonomy, editability, or storage abstraction as specialized memory banks. For this reason, we treat them as informative related mechanisms rather than core representatives in the main taxonomy table.

\paragraph{Conditional and Scalable Lookup Memory}
More recent work revisits lookup-based memory within model-integrated architectures. Engram~\cite{cheng2026conditionalmemoryscalablelookup} introduces a scalable conditional memory mechanism in which information is stored in sparsely activated slots and retrieved via structured hashing. Rather than recomputing historical context through full attention, the model performs selective memory lookup conditioned on the current hidden state. The memory slots constitute persistent storage components whose contents are not automatically discarded after inference. Importantly, although Engram integrates tightly with the forward pass, the stored memory representations remain structurally distinct from transient activations. Engram is particularly informative because it separates memory sparsity from expert sparsity: MoE layers sparsely activate parameterized transformations, whereas Engram sparsely addresses stored memory entries. This distinction clarifies why lookup memory should not be reduced to conditional computation alone. Its central contribution is the introduction of a scalable addressing substrate whose capacity can grow independently from dense computation, subject to the accuracy and collision behavior of the indexing scheme.

\paragraph{Editable Memory Banks}
Another line of work makes the stored knowledge itself more explicit and editable. PlugLM~\cite{cheng2023decoupleknowledgeparameters} replaces selected feed-forward layers with a differentiable plug-in key--value memory, motivated by the observation that FFNs already behave like implicit key--value stores. Its memory entries can be retrieved through knowledge attention and updated without full model retraining, which places it between neural parameter memory and external lookup memory. ExplicitLM~\cite{yu2025explicitlmdecouplingknowledgeparameters} strengthens this separation by introducing layer-wise explicit memory banks whose entries are human-readable token sequences and are accessed through a two-stage differentiable retrieval mechanism. These works are not simply RAG systems, because the memory banks are trained and invoked as architectural components rather than appended through an external orchestration pipeline. Their significance for explicit memory is twofold: they make stored knowledge more inspectable than ordinary FFN parameters, and they expose targeted update operations that are difficult to realize in entangled parametric memory.

\paragraph{Feed-Forward Layers as Lookup Memory}
MemoryLLM~\cite{jaiswal2026memoryllmplugnplayinterpretablefeedforward} provides another perspective on model-integrated lookup memory by reinterpreting feed-forward networks as context-free token-wise retrieval modules. Instead of treating FFNs only as opaque nonlinear transformations following attention, it trains decoupled FFN memories directly from token embeddings, enabling their outputs to be precomputed as token-level lookups. This work is conceptually different from online self-updatable MEMORYLLM~\cite{wang2024memoryllmselfupdatablelargelanguage}: the former emphasizes interpretable and potentially offloadable feed-forward memory, while the latter introduces a latent memory pool refreshed during inference. From the standpoint of explicit memory, the plug-and-play MemoryLLM weakens the dependence of FFN memory access on the surrounding attention context and turns part of the Transformer into a more explicit storage-and-lookup substrate. This makes it a useful bridge between parameter memory and retrieval memory: the stored information remains in neural parameters, but its access pattern is made closer to a structured lookup table than to ordinary dense computation.

\paragraph{Distinguishing Model-Level Retrieval from RAG}
It is important to clarify the boundary between lookup-based explicit memory and retrieval-augmented generation (RAG). RAG systems typically rely on external document corpora, vector databases, and orchestration pipelines that exist outside the model architecture. In contrast, the mechanisms discussed here embed persistent storage directly into the model’s structural design. The memory repository—whether implemented as a matrix, datastore, or slot-based structure—is treated as an architectural component rather than an external engineering layer. Our focus is strictly on model-level persistent storage.

\paragraph{Structural Properties}
Lookup-based explicit memory systems share several defining characteristics. First, memory entries are individually addressable rather than compressed into a single evolving hidden state. Second, storage capacity scales with the number of slots or datastore size, rather than being bounded by fixed hidden dimensions. Third, retrieval is typically content-based, introducing sparsity and conditional computation. These properties distinguish lookup memory from both attention-based implicit memory (which recomputes context each time) and parameterized memory modules (which encode information into weight tensors). From the update-rule perspective, many lookup systems rely less on online parametric writing and more on structural or admission-based updates: the memory table may be constructed offline, while inference-time dynamics determine which entries become accessible. Collectively, lookup-based memory mechanisms offer a pathway toward scalable and persistent knowledge storage in large language models. By decoupling storage from transient computation and enabling selective addressing, they extend the temporal horizon of model behavior beyond the computation-bound limits. However, they also introduce challenges in memory growth, indexing efficiency, and interference management.

\subsection{Conditional Parameter Memory and Mixture-of-Experts}
\label{subsec:moe_memory}

Mixture-of-Experts (MoE) architectures introduce a distinct form of offline explicit memory embedded in parameter space. Unlike dense Transformer layers—where all parameters participate in every forward pass—MoE models maintain a collection of specialized expert subnetworks and employ a learned routing function to activate only a sparse subset conditioned on the input. Formally, given hidden state $h$, a router produces a sparse gating distribution over $E$ experts:
\begin{equation}
    y = \sum_{i \in \mathcal{S}(h)} g_i(h) \cdot \text{Expert}_i(h),
\end{equation}
where $\mathcal{S}(h)$ is a small selected subset, and $g_i(h)$ is the gating weight. This conditional computation paradigm was popularized by Switch Transformer \cite{fedus2021switch} and GLaM \cite{du2022glam}, demonstrating that sparse expert activation can scale model capacity to trillions of parameters while keeping per-token FLOPs nearly constant. From a memory perspective, each expert constitutes a persistent parameterized memory block. The routing network performs context-dependent addressing over this repository. Unlike retrieval based mechanisms, which access external key-value stores, MoE retrieves transformations directly from parameter space. Nevertheless, both share a structural principle: \emph{explicit, conditional access to modular knowledge storage}. Recent large-scale open models further highlight this interpretation. Mixtral employs top-2 routing over eight experts per layer, achieving strong performance-to-compute tradeoffs while exhibiting emergent expert specialization~\cite{mixtral2024}. DeepSeek-MoE introduces fine-grained expert partitioning and load-balancing strategies to enhance specialization stability and efficiency~\cite{deepseek2024moe}. Empirical analyses in these models show that experts tend to cluster around linguistic, reasoning, or domain-specific subfunctions, reinforcing the view that MoE implements modularized persistent knowledge memory. Within our taxonomy, MoE occupies an distinct intermediate position between explicit parameterized and lookup memory. It transforms parameter space into an addressable memory system governed by learned routing, exemplifying persistent, selectively accessed, and structurally modular memory systems.

\subsection{Multi-Timescale and Nested Update Mechanisms}
\label{subsec:multi-timescale}
A defining advantage of various explicit memory systems lies in their flexibility of temporal evolution. Because memory parameters are structurally distinct from backbone weights, their update schedules need not strictly coincide with standard training or inference steps. This enables multi-timescale learning dynamics, in which different memory parameter subsets evolve at distinct temporal frequencies. Rather than treating memory as a static store, these approaches carefully regulate \emph{how} and \emph{when} memory is modified.

\paragraph{Test-Time Adaptation as Short-Timescale Memory}
Test-time learning mechanisms such as Titans~\cite{behrouz2025titans} and end-to-end test-time training (TTT-E2E)~\cite{tandon2025endtoend} demonstrate that memory parameters can be updated during inference for adaptation. In Titans, a dedicated memory module is adjusted token-by-token using surprise-driven signals, while TTT-E2E introduces differentiable test-time objectives optimized jointly with the backbone. In both cases, memory updates occur at a much higher frequency than pretraining updates, allowing rapid adaptation to local context distributions. From a temporal perspective, such mechanisms instantiate \emph{short-timescale explicit memory}: persistent parameters that evolve dynamically within an inference episode.

\paragraph{Nested and Hierarchical Update Schedules}
Nested Learning~\cite{behrouz2025nested} generalizes this idea by introducing hierarchically organized update frequencies. Different parameter subsets operate under distinct learning rates and update schedules, creating a layered structure of fast-adapting and slow-adapting components. Rather than a binary distinction between backbone and memory, nested update mechanisms form a spectrum of temporal plasticity. 

\paragraph{Slow–Fast Parameter Decomposition}
The notion of multi-timescale learning connects to classical fast–slow weight decompositions~\cite{schmidhuber1992learning,ba2016using}, where rapidly updated parameters capture recent patterns while slowly updated weights encode stable knowledge. In large language models, this decomposition re-emerges as a practical strategy for balancing stability and adaptability. By restricting high-frequency updates to a limited parameter subset, models reduce the risk of catastrophic interference while retaining the ability to accumulate contextual information. Importantly, this separation is architectural rather than incidental: the memory parameters are explicitly designated for temporal adaptation.


\paragraph{Implications for Stability–Plasticity Tradeoffs}
Multi-timescale mechanisms directly engage the classical stability–plasticity dilemma: how to incorporate new information without overwriting previously stored knowledge. Frequent updates increase adaptability but risk overfitting to transient signals, whereas slower updates preserve stability at the cost of responsiveness. Explicit online memory architectures must therefore regulate update magnitude, frequency, and scope. Nested and hierarchical update strategies offer one principled avenue for navigating this tradeoff, highlighting that temporal control is as central to explicit memory as structural persistence. In summary, multi-timescale update mechanisms extend explicit memory beyond static parameter repositories. They reveal that persistence alone may be insufficient to characterize memory behavior; the temporal dynamics of parameter evolution constitute an equally critical dimension. By decoupling storage from computation and enabling flexible update schedules, explicit memory systems open a broader design space for long-horizon adaptation in LLMs.

\subsection{Structural Implications and Risks}

While explicit and persistent memory systems expand the temporal adaptability of large language models, they may also introduce structural complexities absent from purely implicit systems. Decoupling storage from transient computation enlarges the design space, but simultaneously creates new challenges in stability, scalability, and controllability.

\paragraph{Capacity Growth and Memory Scaling}
Unlike implicit memory, whose capacity is bounded by hidden-state dimensionality and context windows, explicit memory may scale with the number of parameters, slots, or datastore entries. Lookup-based systems can grow unbounded as new entries accumulate, and parameterized memory modules may expand to accommodate richer representations. However, increased capacity also raises computational and storage costs, and may introduce diminishing returns as memory grows. Efficient indexing, pruning strategies, and compression mechanisms become necessary to prevent unmanageable expansion.

\paragraph{Interference and Memory Drift}
Persistent but online storage implies that newly written information coexists with previously stored content. Without careful regulation, memory updates can lead to interference, overwriting useful signals or amplifying noise. In parameterized modules updated at test time, rapid adaptation risks destabilizing learned representations. In lookup-based systems, stale or redundant entries may bias retrieval. Multi-timescale update strategies partially mitigate these effects, yet the stability–plasticity tradeoff remains a central unresolved tension.

\paragraph{Optimization and Convergence Challenges}
Explicit memory systems often introduce additional objectives or update loops, such as surprise-driven learning rules or auxiliary loss functions. These mechanisms may not share the same optimization guarantees as standard pretraining. Frequent test-time updates can accumulate bias, while nested update hierarchies complicate convergence analysis. Unlike implicit memory—whose dynamics are fully captured by forward computation—explicit memory requires reasoning about coupled learning processes operating at multiple temporal scales.

\paragraph{Consistency Across Episodes}
Because explicit memory persists beyond single inference calls, questions of memory lifecycle management arise. When should memory be reset, how should long-term and short-term storage be separated, and what mechanisms ensure that accumulated information remains relevant, these questions resemble those studied in continual learning, suggesting deeper connections between explicit memory design and lifelong learning paradigms.

\paragraph{Architectural Complexity and Interpretability}
Introducing persistent storage modules increases architectural heterogeneity. Models now contain backbone weights, memory parameters, retrieval indices, and update controllers. While this separation enhances controllability, it complicates interpretability and system-level analysis. Understanding which component stores what information—and how updates propagate through the model remains an open research challenge.

\paragraph{Geometric Constraints for Parametric Stability}
Although this survey does not treat continual-learning optimizers as a primary memory architecture, recent constrained adaptation methods are relevant to the stability of long-term parametric memory. Orthogonal low-rank adaptation in Lie groups~\cite{cao2026orthogonallowrankadaptationlie} and Muon-OGD~\cite{lu2026muonogdmuonbasedspectralorthogonal} both aim to reduce interference when parameters are updated across tasks, respectively by constraining low-rank updates on structured manifolds or by imposing spectral-norm-aware orthogonal gradient projections. These methods are not explicit memory modules in the same sense as Titans, Engram, or LM2, because they primarily regulate how existing parameters are adapted. Nevertheless, they point to an important open issue for memory-centric LLMs: as more architectures introduce writable parameter subsets, the geometry of the update rule may become as important as the storage form itself for preventing drift and preserving previously encoded information.

\paragraph{Comparison with Implicit Memory}
It is instructive to contrast these risks with those of implicit memory. Implicit mechanisms are often limited by computation-bound capacity and context length, but benefit from architectural simplicity and relatively well-understood optimization dynamics. Explicit memory removes the former constraints at the cost of additional structural and training complexity. Thus, the choice between implicit and explicit memory is not merely a matter of capacity, but of system design philosophy: tightly coupled computation dynamics versus modular persistent storage. In summary, explicit memory mechanisms extend the functional horizon of large language models, enabling adaptation and persistence beyond the scope of transient computation. However, this expansion introduces fundamental tradeoffs in scalability, optimization stability, and interference control. Addressing and analyzing these structural challenges is essential for realizing the full potential of explicit memory architectures.

\section{Model-Level Memory Architectures: Design, Implementation, and Evaluation}
\label{sec:systems}

Many successful language models rely on a comparatively simple memory paradigm—most notably, attention-based working memory in standard Transformer architectures. Despite their simplicity, such models have demonstrated remarkable performance across a wide range of tasks. However, recent architectural developments increasingly suggest that diversifying and integrating multiple memory mechanisms can yield further gains in efficiency, scalability, and long-term capability for practical systems. Rather than relying on a single computational memory pathway, emerging models explore structured combinations of implicit and explicit memory components, multi-timescale updates, and specialized memory modules. This trend motivates an integrative model-architecture perspective: how different memory mechanisms are composed within a unified model, how they are supported and optimized in practice, and how their memory capabilities are evaluated. In this section, we examine memory from this viewpoint. We first analyze hybrid memory architectures that combine heterogeneous memory paradigms (\S\ref{subsec:hybrid_systems}). We then discuss architectural-level memory management and efficiency considerations (\S\ref{subsec:memory_management}), followed by a review of evaluation frameworks that measure diverse memory capabilities (\S\ref{subsec:evaluation}).

\subsection{Hybrid Memory Architectures}
\label{subsec:hybrid_systems}

Section~\ref{sec:implicit} analyzed hybridization at the level of implicit memory mechanisms—such as interleaving attention and state dynamics within a single representation. In contrast, this section focuses on \emph{architectural hybridization} of memory paradigms in model design: how distinct memory mechanisms are combined within a unified model to leverage complementary strengths without relying on external systems.

\paragraph{Interleaved Attention and Structured State Layers.}
A prominent pattern in recent research is the hybridization of attention-based and state-space based sequence models. For example, Samba~\cite{ren2024samba} interleaves Mamba-style selective state space layers with sliding-window attention blocks, enabling efficient handling of ultra-long contexts while preserving high-resolution recall in recent tokens. Trained at scale, Samba demonstrates strong performance and extrapolation behavior across long-context benchmarks, surpassing pure Transformer and pure SSM baselines on both perplexity and memory recall tasks. Similarly, LightTransfer~\cite{zhang2024light} proposes transforming a pretrained Transformer into a hybrid model by identifying and replacing ``lazy'' layers with streaming-optimized components, trading off local attention computation for lightweight recurrent or streaming mechanisms with minimal fine-tuning. This approach highlights a different axis of hybrid design: \emph{layer-wise adaptation} of memory mechanism depending on locality and computational role. Priming~\cite{chattopadhyay2026priminghybridstatespace} extends this direction by treating hybridization as a knowledge-transfer problem from pretrained Transformers to hybrid state-space architectures. Rather than defining a new memory representation, it selects attention layers for replacement, aligns the substituted state-space modules with the original layer behavior, and then performs post-training. Its relevance here is methodological: it lowers the cost of moving from attention-dominated working memory toward mixed attention--recurrent memory without requiring full pretraining from scratch.

\paragraph{Adaptive Hybrid Memory Routing.}
Recent hybrid architectures increasingly move beyond choosing a fixed ratio of attention and recurrent layers. Fixed serial hybrids and hybrid-head models, including OLMo Hybrid~\cite{merrill2026olmohybridtheorypractice}, Kimi Linear~\cite{kimiteam2025kimilinearexpressiveefficient}, Hymba~\cite{dong2024hymbahybridheadarchitecturesmall}, and Falcon-H1~\cite{zuo2025falconh1familyhybridheadlanguage}, demonstrate that attention and recurrent/state-space memory can be combined at different depths or within the same layer. However, these designs generally allocate the high-fidelity attention pathway according to static architectural choices: attention is placed in fixed layers, fixed heads, or fixed channel partitions, and the KV cache still tends to grow with all tokens processed by attention. Recent analyses of hybrid linear attention further suggest that the precise allocation of full attention is a central determinant of the recall--efficiency tradeoff~\cite{wang2026systematicanalysishybridlinear}. AMOR~\cite{zheng2026thinkfastslowamor} introduces a more adaptive alternative by appending post-hoc attention refinement blocks to a recurrent backbone and activating them only for positions with high normalized output entropy. In this design, the recurrent backbone remains a complete language model, while attention is used as a conditional refinement mechanism for uncertain or information-dense tokens. HAM~\cite{lufkin2026hybridassociativememories} makes the complementary-memory principle more explicit inside the sequence-mixing layer: a recurrent memory processes every token, whereas a sparse KV cache stores only tokens that are difficult for the recurrent state to predict, with the cache growth controlled by a threshold or learned router. A related fixed-budget precursor blends quadratic and linear memories with a bounded KV component~\cite{irie2025blendingcomplementarymemorysystems}; HAM differs by making cache admission data-dependent and adjustable at runtime. Together, these models suggest an emerging direction in hybrid memory design: the key question is not only where attention should be inserted, but also which tokens should receive high-resolution storage or attention-based refinement.

This line of work is especially useful for interpreting update rules in hybrid architectures. Fixed serial or hybrid-head models mainly use scheduled or structural allocation: the architecture decides in advance which layers or heads maintain high-resolution attention memory. AMOR and HAM instead introduce signal-gated updates over memory access. AMOR gates whether attention refinement is executed, while HAM gates whether a token is admitted into the KV cache. In both cases, the stored or accessible memory set is updated according to uncertainty or prediction error, making hybrid memory a dynamic resource rather than a static layer layout.

\paragraph{Implicit Memory with Explicit Storage.}
More structurally significant are systems that combine computation-based memory with explicit storage mechanisms. Titans~\cite{behrouz2025titans}, for instance, augments a Transformer backbone with a test-time learnable memory module that updates parameters during inference. The Transformer provides conventional short-term memory, while the auxiliary module accumulates task-specific information across interactions. This separation introduces a dual-memory regime: transient activation-based memory and persistent parameterized explicit memory. Lookup-based augmentations offer another integration pattern. Systems inspired by conditional memory modules such as Engram~\cite{cheng2026conditionalmemoryscalablelookup} incorporate scalable key-based retrieval mechanisms within the model’s early layers, effectively routing tokens through sparse memory slots. In such architectures, explicit memory storage coexists with dense Transformer computation, allowing selective recall without uniformly increasing computational cost.

\paragraph{Multi-Timescale Hybridization.}
Hybrid systems often differ not only in representation but also in update timescale. For example, End-to-End Test-Time Training (TTT-E2E)~\cite{tandon2025endtoend} performs batch-style parameter updates during inference, whereas Nested Learning~\cite{behrouz2025nested} decomposes learning into multiple nested optimization loops operating at distinct frequencies. When integrated into larger architectures, these mechanisms create memory hierarchies in which activation states evolve at token-level granularity, explicit memory modules update at interaction or batch level, and backbone parameters remain frozen or adapt slowly.

\paragraph{Multi-Component and Modular Hybrid Designs.}
Beyond simple layer interleaving, some architectures explicitly integrate multiple memory-relevant components to address varied modeling demands. Hydra~\cite{chaudhary2025hydramodulararchitectureefficient}, for example, combines structured state space backbones with sparse global attention, mixture-of-experts (MoE) feed-forward routing, and dual workspace and factual memory mechanisms. This modularization illustrates how distinct components—each with different memory characteristics—can be orchestrated to balance long-range context propagation, sparse access efficiency, and specialized computation. Another line of work explores hybrid memory span augmentation. Expansion Span~\cite{nunez2025expansion} introduces span-expanded attention in SSM–attention hybrids, reserving portions of the attention context for retrieved tokens beyond the usual finite context window. By combining conditional retrieval with state propagation and local attention, this mechanism extends the effective eidetic memory span of hybrid models without incurring full cost of large context windows.

\paragraph{Design Tradeoffs within Hybrid Models.}
Hybrid memory architectures typically aim to reconcile distinct desiderata:

\begin{itemize}
    \item \textbf{High-Resolution Recall vs. Global Context:} Attention layers provide detailed local associations, while state dynamics or sparsified components aggregate distant dependencies in a compressed form.
    \item \textbf{Computational Efficiency:} Structured state or streaming layers often achieve linear time and space complexity compared to quadratic attention, enabling scalability to longer contexts without prohibitive costs.
    \item \textbf{Modularity and Reuse:} Modular memory components, such as MoE or hybrid attention heads, facilitate selective activation and specialization, controlling when and how different memory pathways are invoked.
\end{itemize}

By organizing multiple memory mechanisms within a single architectural blueprint, hybrid designs demonstrate that the space between isolated memory paradigms can be productively explored. Various architectures orchestrate multiple complementary mechanisms to achieve broader memory capabilities within model layers, while preserving end-to-end differentiability and training efficiency.

\subsection{Memory Management and Efficiency}
\label{subsec:memory_management}

As memory capacity expands through long context windows, hybrid architectures, and explicit modules, practical deployment of large language models increasingly hinges on efficient memory management. At the system level, this challenge primarily manifests in managing attention caches, reducing memory bandwidth bottlenecks, and implementing structured sparsity mechanisms that preserve modeling fidelity while controlling resource usage.

\paragraph{KV Cache Compression and Quantization.}
In autoregressive Transformers, the KV cache constitutes a growing working memory whose size scales linearly with sequence length. For long-context inference, KV storage often becomes the dominant memory bottleneck. Recent work explores compressing the KV cache while retaining retrieval fidelity. CommVQ proposes vector quantization of communication states, significantly reducing KV storage and inter-device communication cost with minimal performance degradation \cite{li2025commvq}. Other approaches perform low-rank projection or structured pruning of cached keys and values, effectively treating the KV cache as a compressible working memory rather than an exact token-level store. A complementary direction is to cache recurrent memory states rather than per-token keys and values. Memory Caching~\cite{behrouz2026memorycachingrnnsgrowing} stores segment-level snapshots of recurrent hidden memory and aggregates them during later computation, creating a middle ground between fixed-size recurrent states and fully token-level Transformer caches. This perspective broadens memory management beyond Transformer KV tensors and shows that working-memory system design also matters for recurrent architectures.

\paragraph{Memory Virtualization and Paged Attention.}
Beyond compression, system-aware architectural modifications have been introduced to virtualize attention memory. PagedAttention \cite{kwon2023efficientmemorymanagementlarge} restructures the KV cache into pageable memory blocks, enabling efficient allocation and reuse across variable-length sequences. By decoupling logical attention memory from contiguous physical allocation, this method reduces memory fragmentation and supports large-batch long-context inference. Although such techniques are often discussed in the context of inference systems (e.g., vLLM), they directly shape how attentional memory is structured and accessed at the architectural level.

\paragraph{Working-Memory Consolidation.}
Not all KV-cache modifications are purely compressive. Bottlenecked Transformers~\cite{oomerjee2026bottleneckedtransformersperiodickv} introduce periodic processing of cached representations, using an auxiliary bottleneck module to consolidate or rewrite parts of the working memory at structured generation boundaries. From the taxonomy perspective, this remains closer to short-term implicit working memory than to long-term explicit storage, because the memory being manipulated is still the model's internal cache rather than an independently persistent repository. Nevertheless, it highlights an important design pattern: attention memory can be managed not only by dropping, quantizing, or paging entries, but also by periodically transforming them into a more compact or stable representation before subsequent reasoning steps.

\paragraph{Sliding Window versus Sparse Global Attention.}
Structured sparsity in attention offers another dimension of memory control. Sliding-window attention restricts each token to attend only to a fixed local neighborhood, yielding linear complexity while preserving short-range precision. In contrast, sparse global attention mechanisms—such as Longformer \cite{beltagy2020longformer} and BigBird \cite{zaheer2020bigbird}—combine local windows with selected global tokens to maintain theoretical expressivity guarantees. These approaches can be interpreted as controlled memory retention strategies: sliding windows prioritize recency, whereas sparse global links preserve long-distance anchors.

\paragraph{Architectural Tradeoffs.}
Collectively, these techniques illustrate that memory management is not merely an engineering afterthought but an architectural design axis. Compression, virtualization, and structured sparsity reshape how memory is represented, accessed, and scaled. These methods refine the operational realization of memory, enabling extended context modeling without proportionally increasing computational or storage cost.

\subsection{Evaluation of Memory Systems}
\label{subsec:evaluation}

Evaluating memory in large language models is fundamentally more challenging than measuring standard language modeling performance. Unlike perplexity or downstream accuracy, memory is not a directly observable scalar quantity. Instead, it manifests through behaviors such as long-range dependency resolution, recall fidelity, robustness to interference, and persistence across contexts. Consequently, targeted evaluation of memory systems requires carefully designed diagnostic tasks that isolate specific memory capabilities.

\paragraph{Long-Context Retrieval and Recall Benchmarks.}
A primary dimension of memory evaluation concerns retrieval accuracy over long contexts. Synthetic tasks such as Needle-in-a-Haystack (NIAH) place a target fact within a large distractor context and test whether the model can recover it. More comprehensive benchmarks such as LongBench \cite{bai2023longbench} aggregate diverse long-context tasks, including multi-document QA, summarization, and reasoning, to assess effective context utilization. RULER \cite{hsieh2024rulerwhatsrealcontext} and L-Eval \cite{an2024leval} further analyze scaling behavior as context length increases, measuring degradation curves across varying sequence sizes. These evaluations probe the operational limits of model memory, revealing phenomena such as attention dilution and recency bias.

\paragraph{Structured Dependency and Reasoning Tests.}
Beyond raw retrieval, memory must support structured reasoning across distant tokens. Benchmarks such as SCROLLS \cite{shaham2022scrolls} and NarrativeQA-style tasks evaluate whether models can integrate information distributed across long documents. These tasks test not only storage capacity but also the model's ability to maintain coherent intermediate representations across extended contexts. Hybrid and state-space models often demonstrate improved stability in such tasks due to their compressed state propagation mechanisms, though they may trade off fine-grained token-level recall.

\paragraph{Forgetting, Interference, and Stability.}
Explicit memory mechanisms introduce additional evaluation dimensions. When models incorporate retrieval modules, gradient-based updates, or conditional parameter memory, questions arise regarding interference and stability. Key evaluation aspects include memory degradation speed, sensitivity towards distractions, and temporal persistence of newly acquired information. For lookup-based models, recall precision and key-value alignment metrics are often used to quantify explicit memory quality. For parameter-based memory updates, continual learning benchmarks provide complementary evaluation signals.

\paragraph{Efficiency–Performance Tradeoffs.}
Memory evaluation must also account for computational and storage cost. Extended context performance can be artificially inflated by increasing window size, yet such gains may be impractical under real-world memory constraints. Recent benchmarks therefore report performance as a function of context length and memory footprint, emphasizing scaling curves rather than single-point metrics. This is particularly relevant for compressed or paged attention systems, where memory management strategies influence effective recall.

\paragraph{Implicit vs. Explicit Memory Evaluation.}
Importantly, implicit and explicit memory paradigms may have different evaluation protocols. Implicit memory is typically evaluated through long-range dependency tasks, accuracy across increasing context lengths, and stability under noise injection. Explicit persistent memory systems, in contrast, may require evaluation of update consistency, retrieval precision and latency, and robustness to memory growth and interference. Thus, memory evaluation should be multi-dimensional, reflecting not only recall accuracy but also stability, efficiency, and adaptability.

\paragraph{Toward Unified Memory Metrics.}
Despite growing benchmark coverage, a unified metric for memory remains elusive. Future work may benefit from decomposing memory performance into orthogonal axes—capacity, fidelity, persistence, and efficiency—allowing systematic comparison across implicit, explicit, and hybrid architectures. A principled evaluation framework is essential for disentangling architectural improvements from mere increases in context size or parameter count. Only through controlled and standardized diagnostics can the true contribution of memory be rigorously assessed.

\section{Open Challenges and Future Directions}
\label{sec:challenges}

The rapid diversification of memory mechanisms in large language models signals a broader shift in architectural philosophy: memory is no longer an incidental byproduct of scaling, but an explicit design axis. From attention-based working memory to structured state dynamics, conditional parameter memory, and retrieval-based systems, contemporary models and systems increasingly treat memory as a modular and controllable component. Despite substantial progress, several foundational challenges remain unresolved. Addressing these challenges will require deeper theoretical grounding, improved adaptability, tighter hardware integration, and more principled evaluation.

\subsection{Toward a Unified Theory of Memory in LLMs}

Current memory mechanisms are often categorized operationally—attention, recurrence, retrieval, parametric—without a unifying formal theoretical abstraction. However, these mechanisms can be interpreted within a broader state-transition perspective, where memory differs primarily in persistence, accessibility, update dynamics, and representational compression. A rigorous theoretical framework could clarify the deeper relationships between various memory paradigms. Such a framework would characterize memory capacity not merely in terms of context length or parameter count, but through measurable properties such as retention fidelity, interference robustness, and information compression efficiency. Formalizing the stability–plasticity tradeoff in dynamic update mechanisms may also illuminate why certain architectures degrade under long contexts or continual adaptation. Developing such theoretical foundations would enable principled comparison across architectures and prevent conflating scale with genuine memory capability.

\subsection{Continual and Lifelong Parametric Memory}

Persistent memory mechanisms that allow online adaptation introduce a new regime of architectural design. Gradient-based test-time updates, nested parameter hierarchies, and modular memory blocks blur the distinction between pretraining and inference. However, these mechanisms also reintroduce classical challenges of continual learning, including forgetting and instability under distribution shift. Future research is likely to focus on advanced stable update protocols, modular isolation strategies, and meta-learned consolidation mechanisms. Rather than treating adaptation as a monolithic parameter update, architectures may increasingly partition memory into components with differentiated plasticity. Such designs would allow models to accumulate knowledge incrementally while preserving previously acquired competencies. Achieving scalable lifelong memory without full retraining remains one of the central open problems in large-scale model design.

\subsection{Robust and Interpretable Update Rules}

The refined update-rule taxonomy~\ref{tab:update_rules} exposes a set of mechanism-specific challenges. Optimization-based memory writing offers strong adaptability, but it is vulnerable to objective mismatch, accumulated bias, and catastrophic drift when the test-time objective is only a proxy for downstream utility. State-transition updates are efficient and stable enough for long streams, yet their compressed states can suffer from interference, saturation, or poorly calibrated forgetting. Signal-gated mechanisms promise more selective memory allocation, but their reliability depends on whether surprise, entropy, or prediction error is a faithful indicator of future relevance; information that appears unimportant locally may become crucial later. Admission, eviction, and consolidation rules face a related irreversibility problem: once a token or memory entry is dropped, compressed, or overwritten, later retrieval may be impossible. A key future direction is therefore to develop update rules that are not only efficient, but also diagnosable, reversible when necessary, and robust under long-horizon distribution shifts.

\subsection{Adaptive Memory Allocation and Control}

As hybrid architectures integrate multiple memory pathways—attention layers, state space modules, retrieval systems, and expert routing—the question shifts from how to design individual mechanisms to how to coordinate them. Most existing models rely on static compositions, where memory pathways are fixed at design time. This rigidity limits the ability to adapt memory usage to task structure or contextual demands. A promising direction lies in adaptive memory orchestration. Learned controllers may dynamically allocate tokens across memory subsystems, compress context representations based on task requirements, or modulate routing across experts and retrieval modules. Such mechanisms would elevate memory from a structural component to a dynamically regulated resource. The transition from fixed hybridization to adaptive coordination represents a natural progression in the architectural evolution of memory systems.

\subsection{Hardware–Algorithm Co-Design for Scalable Memory}

As context windows extend to hundreds of thousands or millions of tokens, memory bandwidth and storage hierarchies increasingly dominate system performance. Architectural innovation can no longer be decoupled from hardware constraints. Future models may integrate hierarchical memory designs aligned with GPU and accelerator memory tiers, enabling multi-level caching and selective recomputation. In-memory attention computation and hardware-aware sparsity patterns suggest that memory operations themselves may become primary computational primitives. Such co-design efforts will be essential for sustaining scalability without prohibitive energy or latency costs. The long-term trajectory of memory architectures will likely be shaped as much by hardware considerations as by modeling objectives.

\subsection{Principled and Multi-Dimensional Memory Evaluation}

Despite expanding benchmark coverage, evaluation of memory systems remains fragmented and not targeted. Most current protocols measure performance under extended context lengths, yet this conflates memory capacity with reasoning ability and other scaling effects. A more principled evaluation paradigm would decompose memory performance into orthogonal dimensions, including storage capacity, retrieval fidelity, temporal persistence, robustness to interference, and computational efficiency. Standardized stress tests—such as controlled interference injection, memory decay measurement, and update consistency analysis—would provide clearer diagnostics of architectural behavior. Establishing such evaluation standards is critical for disentangling genuine memory innovations from improvements driven by conventional scaling or dataset overlap. Only with multi-dimensional and controlled diagnostics can architectural advances in memory be rigorously assessed.



\section{Conclusion}

This survey has formalized a mechanism-centric perspective on memory in large language models (LLMs), synthesizing a broad spectrum of architectural innovations into a unified design space. We proposed a taxonomy structured along three orthogonal axes: representation (implicit vs.\ explicit), update dynamics (offline vs.\ online), and persistence (short-term vs.\ long-term). By decomposing update dynamics into optimization-based writing, state transitions, signal-gated routing, and structural consolidation, our framework precisely characterizes how information is retained, modified, and forgotten within modern architectures.

Our analysis reveals that while implicit mechanisms---such as attention and recurrent sequence memory---are tightly coupled with forward computation, explicit memory introduces autonomous, persistent storage via writable parameters and addressable lookups. We highlight that no single memory paradigm satisfies all computational desiderata. Token-level attention ensures high-fidelity recall at the expense of scalability; recurrent sequence memory offers compressed, long-horizon views but faces interference; and explicit memory enhances temporal persistence while introducing complexities in update stability and addressing. Consequently, modern LLM design is rapidly shifting toward hybrid architectures that adaptively orchestrate these complementary substrates.

Ultimately, memory in LLMs has transitioned from a byproduct of scale to a first-class compositional design problem. Advancing this frontier necessitates rigorous theoretical foundations, robust and interpretable update rules, adaptive memory routing, hardware-algorithm co-design, and multi-dimensional evaluation protocols. By providing a principled vocabulary and structural framework, this survey aims to catalyze the development of next-generation language models endowed with scalable, robust, and controllable memory architectures.


\bibliographystyle{TPAMI/IEEEtran}
\bibliography{TPAMI/IEEEabrv,TPAMI/ref}


 




\vfill

\end{document}